\documentclass[sigconf]{acmart}

\AtBeginDocument{%
  \providecommand\BibTeX{{%
    \normalfont B\kern-0.5em{\scshape i\kern-0.25em b}\kern-0.8em\TeX}}}

\setcopyright{acmlicensed}
\copyrightyear{2022}
\acmYear{2022}
\acmConference[WWW '22] {Proceedings of the ACM Web Conference 2022}{April 25--29, 2022}{Virtual Event, Lyon, France.}
\acmBooktitle{Proceedings of the ACM Web Conference 2022 (WWW '22), April 25--29, 2022, Virtual Event, Lyon, France}
\acmPrice{15.00}
\acmISBN{978-1-4503-9096-5/22/04} 
\acmDOI{10.1145/3485447.3511945}

\usepackage{multirow}
\usepackage{booktabs}
\usepackage{natbib}
\usepackage[ruled,vlined]{algorithm2e}
\usepackage[utf8]{inputenc}
\usepackage{amssymb,amsmath,caption}
\usepackage[hang,flushmargin]{footmisc}
\usepackage{lipsum}
\usepackage{listings}
\usepackage{extarrows}
\usepackage{subfigure}
\usepackage{xspace}
\usepackage{graphicx}
\usepackage{makecell}
\usepackage{ifthen}
\usepackage{xifthen}
\usepackage{wrapfig}
\usepackage{amsthm}
\usepackage{paralist} 

\theoremstyle{definition}

\newtheorem{theorem}{Theorem}
\newtheorem{proposition}{Proposition}

\newenvironment{pf}{{ \it{Proof.}}\quad}{}

\newcommand{\solution}[0]{SelfKG\xspace}

\newcommand{\vpara}[1]{\vspace{0.07in}\noindent\textbf{#1}\xspace}

\newcommand{\beqn}[1]{{\small\begin{eqnarray}#1\end{eqnarray}}}

\newcommand{\trieq}[0]{\triangleq}
\newcommand*{\iidsim}{\overset{\text{i.i.d.}}{\sim}}
\newcommand{\hide}[1]{} 

\newcommand{\hhy}[1]{{{#1}}} 

\DeclareDocumentCommand{\distn}{mO{}mO{}O{}}{%
    {#1}\ifthenelse{\isempty{#2}}{}{_{#2}}%
    \ifthenelse{\isempty{#3}}{}{\left[{#3}\ifthenelse{\isempty{#4}}{}{\mathrel{}\middle\vert\mathrel{}#4}\ifthenelse{\isempty{#5}}{}{; {#5}}\right]}
}
\DeclareDocumentCommand{\prob}{O{}mO{}O{}}{%
    \distn{\mathbb{P}}[#1]{#2}[#3][#4]
}
\DeclareDocumentCommand{\expect}{O{}mO{}}{%
    \mathbb{E}\ifthenelse{\isempty{#1}}{}{_{#1}}%
    \ifthenelse{\isempty{#2}}{}{\left[{#2}\ifthenelse{\isempty{#3}}{}{\mathrel{}\middle\vert\mathrel{}#3}\right]}
}
\DeclareDocumentCommand{\expectunder}{O{}mO{}}{%
    \underset{{#1}}{\mathbb{E}}%
    \ifthenelse{\isempty{#2}}{}{\hspace{-3pt}\left[{#2}\ifthenelse{\isempty{#3}}{}{\mathrel{}\middle\vert\mathrel{}#3}\right]}
}
\DeclareDocumentCommand{\expectundernear}{O{}mO{}}{%
    \underset{{#1}}{\mathbb{E}}%
    \ifthenelse{\isempty{#2}}{}{\hspace{-5pt}\left[{#2}\right]}
}
\DeclareDocumentCommand{\var}{O{}mO{}}{%
    \text{Var}\ifthenelse{\isempty{#1}}{}{_{#1}}%
    \ifthenelse{\isempty{#2}}{}{\left[{#2}\ifthenelse{\isempty{#3}}{}{\mathrel{}\middle\vert\mathrel{}#3}\right]}
}
\DeclareDocumentCommand{\indic}{O{}mO{}}{%
    \mathds{1}\ifthenelse{\isempty{#1}}{}{_{#1}}%
    \ifthenelse{\isempty{#2}}{}{\left({#2}\right)}
}

\newcommand*{\distnpos}{\ppos}
\newcommand*{\ppos}{p_\mathsf{pos}}

\newcommand{\T}[0]{^{\mathsf{T}}}

\begin{document}

\title{SelfKG: Self-Supervised Entity Alignment in Knowledge Graphs}

\author[ X. Liu, H. Hong, X. Wang, Z. Chen, E. Kharlamov, Y. Dong, J. Tang]{
    Xiao Liu$^{\dagger}$, Haoyun Hong$^{\dagger}$, Xinghao Wang$^\dagger$, Zeyi Chen$^{\dagger}$, Evgeny Kharlamov$^\ddagger$, \\Yuxiao Dong$^{\dagger}$, Jie Tang$^{\dagger}$
}
\affiliation{
    $^\dagger$ Department of Computer Science and Technology, Tsinghua University\country{China}
    $^\ddagger$ Bosch Center for AI
}
\email{
  {liuxiao21, honghy17, xinghao-18, chenzeyi19}@mails.tsinghua.edu.cn
}
\email{
  evgeny.kharlamov@de.bosch.com,
  {yuxiaod,jietang} @tsinghua.edu.cn
}
  \thanks{Xiao and Haoyun contributed equally to this work. 
  \thanks{Jie Tang is the corresponding author}
  }

\renewcommand{\shortauthors}{X. Liu, H. Hong, X. Wang, Z. Chen, E. Kharlamov, Y. Dong, J. Tang}
\renewcommand{\authors}{Xiao Liu, Haoyun Hong, Xinghao Wang, Zeyi Chen, Evgeny Kharlamov, Yuxiao Dong, Jie Tang}
  



\begin{abstract} \label{sec:abs}
Entity alignment, aiming to identify equivalent entities across different knowledge graphs (KGs), is a fundamental problem for constructing Web-scale KGs. 
Over the course of its development, the label supervision has been considered necessary for accurate alignments. 
Inspired by the recent progress of self-supervised learning, we explore the extent to which we can get rid of supervision for entity alignment. 
Commonly, the label information (positive entity pairs) is used to supervise the process of pulling the aligned entities in each positive pair closer.
However, our theoretical analysis suggests that the learning of entity alignment can actually benefit more from 
pushing unlabeled negative pairs far away from each other than pulling labeled positive pairs close. 
By leveraging this discovery, we develop the self-supervised learning objective for entity alignment.
We present \solution with efficient strategies to optimize this \hhy{objective} for aligning entities without label supervision. 
Extensive experiments on benchmark datasets demonstrate that \solution~without supervision can 
match or achieve comparable results with state-of-the-art supervised baselines. 
The performance of \solution suggests that self-supervised learning offers great potential for entity alignment in KGs. 
{The code and data are available at \url{https://github.com/THUDM/SelfKG}}.
\end{abstract}



%
\maketitle

\section{INTRODUCTION} \label{sec:intr}

Knowledge graphs (KGs) have found widespread adoption in various Web applications, such as search~\cite{eder2012knowledge,paulheim2017knowledge}, recommendation~\cite{guo2020survey,li2020alimekg}, and question answering~\cite{yang2018hotpotqa,lewis2020retrieval}.  
Constructing large-scale KGs has been a very challenging task.
While we can extract new facts from scratch, aligning existing (incomplete) KGs together is practically necessary for real-world application scenarios. 
Over the past years, the problem of entity alignment~\cite{GCN-Align,tang2019bert-int}, or namely ontology mapping~\cite{li2008rimom} and schema matching~\cite{li2014rule}, has been a fundamental problem for \hhy{the Web research community}.

\begin{figure}[t]
\setlength{\abovecaptionskip}{-0.2mm}
\centering   
\includegraphics[width=\columnwidth]{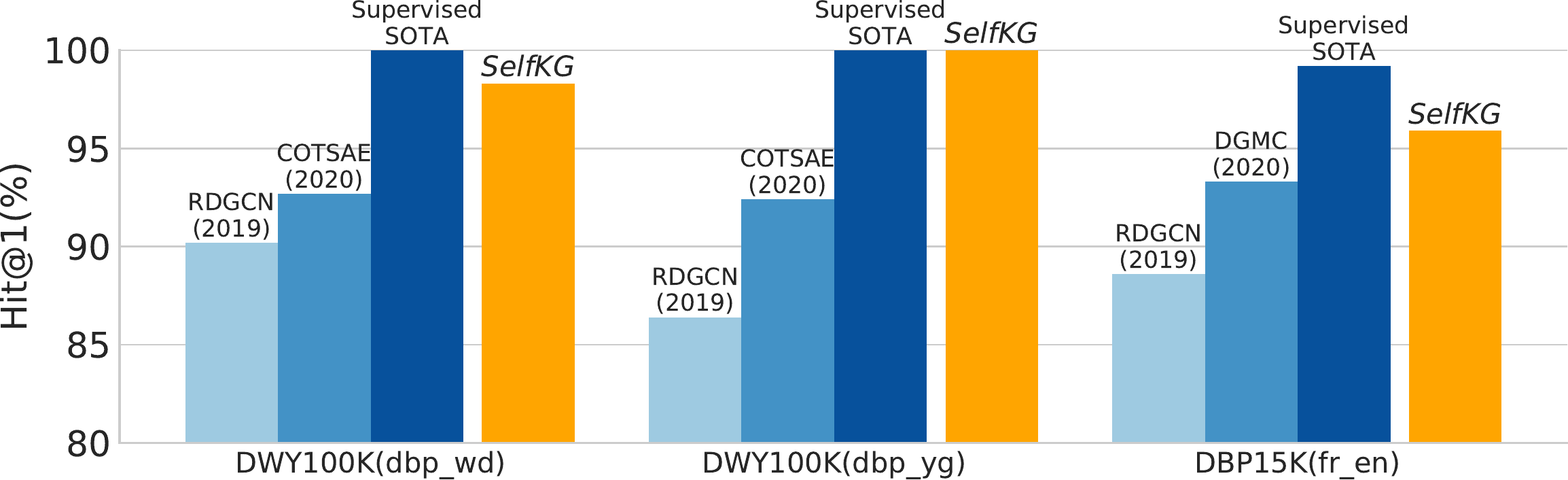}
\vspace{-1.5mm}
\caption{Hit@1 on DWY100K and DBP15K for \solution (0\% of training labels) and SOTA supervised (100\% of training labels) entity alignment. \textmd{Without using any labels, the self-supervised \solution outperforms most of supervised models.}}
\label{fig:intro}
\vspace{-3.5mm}
\end{figure}

Recently, the representation learning-based alignment methods~\cite{MTransE,GCN-Align,CEAFF,tang2019bert-int,wu2019relation} have emerged as the mainstream solutions for entity alignment due to their superior flexibility and accuracy. 
However, their success relies heavily on the supervision provided by human labeling, which can be biased and arduously expensive to obtain for Web-scale KGs. 
In light of this fundamental challenge, we aim to explore the potential to align entities across KGs without label supervision (i.e., self-supervised entity alignment). 

To achieve this, we revisit the common process of the established supervised entity alignment approaches. 
Conceptually, for each paired entities from two KGs, the goal of the existing learning objectives is to make them more similar to each other if they are actually the same entity (i.e., a positive pair), otherwise dissimilar if they are different entities (i.e., a negative pair). 
In the embedding space, this goal is pursued by pulling aligned entities closer and pushing different entities farther away.

We identify the parts where supervision is required in this process. At first place, the supervision serves to pull aligned entities closer.
Secondly, another issue arises is the procedure of generating label-aware negative pairs. 
For every entity in a KG, in the training its negative pairs are formed by randomly sampling entities from the other KG while excluding the groundtruth. 
If without supervision, it is likely that the implicitly aligned entities are sampled as negative pairs, thus spoiling the training (i.e., collision).

\vpara{Contributions.} 
We introduce the problem of self-supervised~\cite{liu2020self} entity alignment in KGs. 
To address it, we present the \solution framework, which does not rely on labeled entity pairs to align entities. 
It consists of three technical components: 1) relative similarity metric, 2) self-negative sampling, and 3) multiple negative queues. 

To get rid of label supervision, we theoretically develop the concept of relative similarity metric (RSM), which enables the self-supervised learning objective. 
The core idea of RSM is that instead of directly pulling the aligned entities closer in the embedding space, it attempts to push not-aligned negatives far away, thus avoiding the usage of the supervision of positive pairs. 
In a relative sense, the (implicitly) aligned entities can be considered to be dragged together when optimizing for RSM. 

By design, to address the dilemma between supervision with label-aware negative sampling and collision of false-negative samples without it, \solution further propose the self-negative sampling strategy, that is, for every entity in a KG, we form its negative pairs by directly sampling entities from the same KG. 
In other words, \solution solely relies on negative entity pairs that are randomly sampled from the input KGs . 
We theoretically show that this strategy remains effective for aligning entities across KGs.

Finally, our theoretical analysis also shows that the self-supervised loss' error term decays faster as the number of negative samples increases, i.e., a large number of negative samples can benefit \solution. 
However, encoding massive negative samples on the fly is computationally very expensive. 
We address this by extending the MoCo technique~\cite{he2020momentum} to support two negative queues, each of which corresponds to the two KGs for alignments, ensuring an efficient increase of negative samples.

Empirically, we conduct extensive experiments to demonstrate the premise of self-supervised entity alignment in KGs. 
We compare the proposed \solution method against 24 supervised and one unsupervised baselines on two widely-used entity alignment benchmarks datasets---DWY100K and DBP15K. 
The results suggest that \solution without using any labels can match or achieve comparable performance with the state-of-the-art supervised baselines (Cf.  Figure \ref{fig:intro}). 
This demonstrates the power of self-supervised learning for entity alignment as well as our design choices of \solution.


\section{PROBLEM DEFINITION}
We introduce the problem of entity alignment in KGs. 
Conceptually, a KG can be represented as a set of triples $T$, each of which denotes the relation $r_{ij}\in R$ between two entities $x_i\in E$ and $x_j\in E$. 
In this work, we denote a KG as $G=\{E,R,T\}$ where $E$, $R$, and $T$ are its entity set, relation set, and triple set, respectively. 

Given two KGs, $G_x = \{E_x,R_x,T_x\}$ and $G_y = \{E_y,R_y,T_y\}$, the set of the existing aligned entity pairs is defined as $S = \{(x,y)|x\in E_x, y \in E_y, x\Leftrightarrow y\}$, where $\Leftrightarrow$ represents equivalence. 
The goal of entity alignment between $G_x$ and $G_y$ is to find the equivalent entity from $E_x$ for each entity in $E_y$, if existed.


 Recently, a significant line of work has been focusing on embedding-based techniques for aligning entities in the vector space, e.g., training a neural encoder $f$ to project each entity $x \in E$ into a latent space. 
Among these attempts, most of them focus on the (semi-) supervised setting in the sense that part of $S$ is used for training the alignment models~\cite{MTransE,GCN-Align,CEAFF,tang2019bert-int,wu2019relation}. 
Due to the limited alignment labels across KGs in the real world, we instead propose to study to what extent the entity alignment task can be solved in an unsupervised or self-supervised setting, under which none of the existing alignments in $S$ is available.


\section{{SELF-SUPERVISED ENTITY ALIGNMENT}} \label{sec:model}

In this section, we discuss the role that the supervision plays in entity alignment and then present the strategies that can help align entities without label supervision. 
To this end, we present the \solution framework for self-supervised entity alignment across KGs.

\subsection{The \solution Framework} \label{sec:architecture}

To enable learning without label information, the main goal of \solution is to design a self-supervised  objective that can guide its learning process. 
To achieve this, we propose the concept of \textit{relative similarity metric} (Cf. Section ~\ref{sec:rsm}) between entities across two KGs. 
To further improve the self-supervised optimization of \solution, we introduce the techniques of \textit{self-negative sampling} (Cf. Section ~\ref{sec:sns}) and \textit{multiple negative queues} (Cf. Section ~\ref{sec:mnq}).


Next, we introduce the initialization of entity embeddings in \solution, which is largely built upon existing techniques, including the uni-space learning and GNN based neighborhood aggregator. 

\vpara{Uni-space learning.} 
The idea of uni-space learning has been adopted by recent (semi-) supervised entity alignment techniques~\cite{{MTransE,GCN-Align,CEAFF,tang2019bert-int,wu2019relation}}. 
Herein, we present how we leverage it for supporting \solution's self-supervised learning setting.

Straightforwardly, embedding entities from different KGs into a uni-space can greatly benefit the alignment task. 
With labeled entity pairs, it is natural to leverage supervision to align different spaces into one, e.g., merging aligned entities for training ~\cite{hao2016joint}, or learning projection matrices with abundant training labels to project entities from different embedding spaces into a uni-space~\cite{MTransE,JAPE}. 

In terms of multi-lingual datasets (e.g., DBP15K), the issue is more challenging. 
Thanks to the pre-trained language models~\cite{han2021pre}, high-quality multi-lingual initial embeddings are now available. For example, the multi-lingual BERT has been used in recent work~\cite{zhang2019multi,tang2019bert-int}. In \solution, we adopt LaBSE~\cite{feng2020language}---a state-of-the-art multi-lingual pre-trained language model trained on 109 different languages---for embedding different knowledge graphs into a uni-space.

\vpara{Neighborhood aggregator. \label{na}}
To further improve the entity embeddings, 
the neighborhood aggregation is used to aggregate neighbor entities' information to the center entity~\cite{GCN-Align,xu2019cross-lingual}. 
In this work, we directly use a single-head graph attention network~\cite{velivckovic2017graph} with one layer to aggregate pre-trained embeddings of one-hop neighbors.

Note that leveraging multi-hop graph structures has been recently explored for the problem of entity alignment. 
Though some studies~\cite{fey2020deep,wu2019relation,GCN-Align} claim that they benefit from multi-hop neighbors, other works~\cite{xu2019cross-lingual,zhang2018mego2vec} argue that one-hop neighbors provides enough information for most situations. 
In our ablation study (Cf. Section ~\ref{sec:ablation}), we find that the multi-hop information actually harms the performance of \solution, which is probably resulted from the distant neighbor noises that may be unignorable in a self-supervised setting. 
Therefore, to demonstrate the minimum requirement of self-supervision for entity alignment, we only involve one-hop neighbor entities during the aggregation. 

\begin{figure}[t]
\centering
\includegraphics[width=\linewidth]{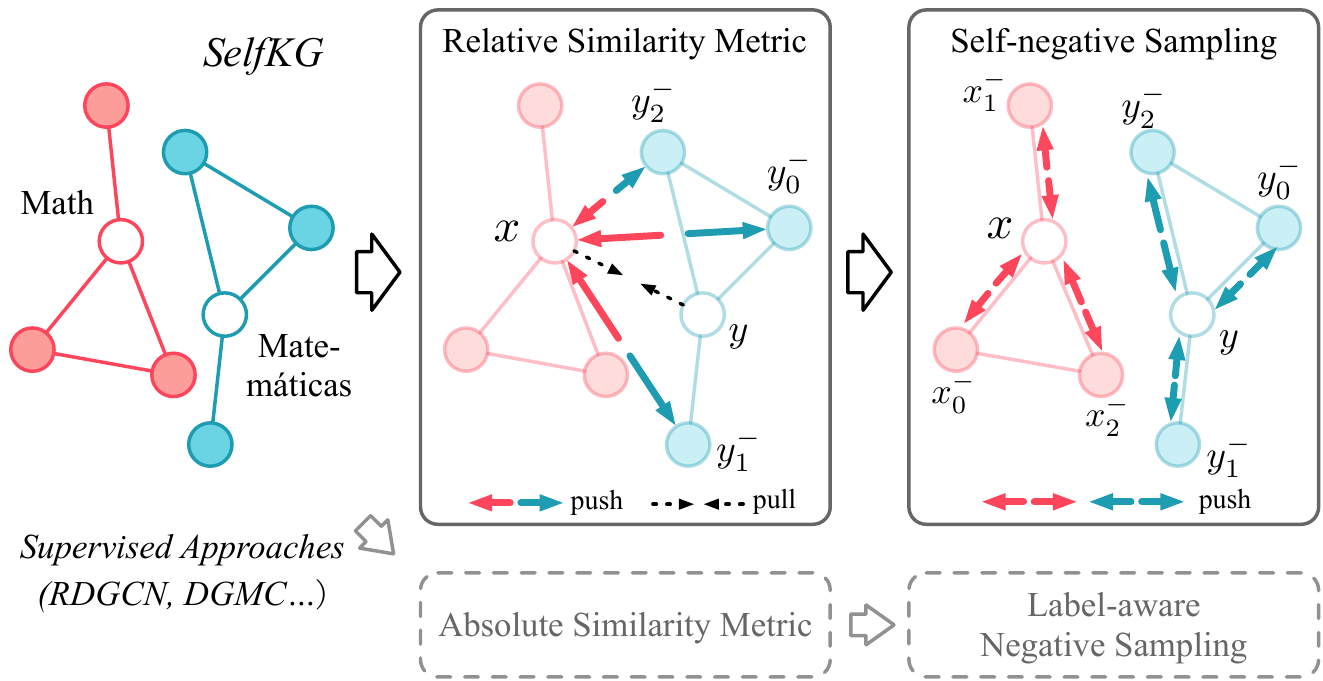}
\caption{A conceptual comparison of \solution and supervised approaches. \textmd{\solution employs the relative similarity metric (RSM) and self-negative sampling to avoid the use of supervision.}
\label{fig:motivations}
}
\vspace{-3mm}
\end{figure}

\subsection{Relative Similarity Metric}
\label{sec:rsm}

We present the self-supervised loss for entity alignment across KGs. 
First, we analyze the supervised NCE loss for entity alignment. 
Then, we introduce the relative similarity metric for avoiding labeled  pairs. 
We finally derive the self-supervised NCE for \solution. 

In representation learning, the margin loss~\cite{bordes2013translating,tang2019bert-int} and cross-entropy loss~\cite{zhang2019oag} have been widely adopted as the similarity metric. 
Without loss of generality, they can be expressed in the form of Noise Contrastive Estimation (NCE)~\cite{gutmann2010noise}. 

In the context of entity alignment, the NCE loss can be formalized as follows. 
Let $p_{\mathsf x}, p_{\mathsf y}$ be the distributions of two KGs $G_\mathrm{x}$, $G_\mathrm{y}$, and $p_{pos}$ denote the representation distribution of the positive entity pairs $(x,y)\in\mathbb{R}^n\times\mathbb{R}^n$.
Given a pair of aligned entities $(x, y) \sim \distnpos$, negative samples ${\{y^-_i\}}_{i=1}^M \iidsim p_{\mathsf y}$, the temperature $\tau$, and the encoder $f$ satisfies $\|f(\cdot)\|=1$, we have the supervised NCE loss as  
\beqn{\scriptsize
\begin{aligned}
\label{eq:nce}
\mathcal{L}_{\rm NCE} &\trieq 
{- \log \frac{e^{f(x)\T f(y) / \tau}}{e^{f(x)\T f(y) / \tau} + \sum_i e^{f(x)\T  f(y^-_i)/ \tau}}}\\
&= \underbrace{-\frac{1}{\tau}f(x)\T f(y)}_{\rm alignment} + \underbrace{\mathop{\log}(e^{f(x)\T f(y) / \tau} + \sum_ie^{f(x)\T f(y^-_i) / \tau})}_{\rm uniformity}.
\end{aligned}
}

\vspace{-0.2cm}
\noindent where the ``alignment'' term is to draw the positive pair close and the ``uniformity'' term is to push the negative pairs away. 

We illustrate how this NCE loss can be further adjusted for a self-supervised setting. 
An example of ``pulling'' and ``pushing'' entity pairs in KGs can be found in Figure ~\ref{fig:motivations} (left). 
Previous studies have shown that the NCE loss has the following asymptotic properties: 

\begin{theorem}{\bf (Absolute similarity metric (ASM)~\cite{wang2020understanding})} \label{th:asm}
For a fixed $\tau > 0$, as the number of negative samples $M \rightarrow \infty$, the (normalized) contrastive loss $\mathcal{L}_{\rm NCE}$ (i.e., $\mathcal{L}_{\rm ASM}$) converges to its limit with an absolute deviation decaying in $\mathcal{O}(M^{-2/3})$. If a perfectly-uniform encoder $f$ exists, it forms the exact minimizer of the uniformity term.
\end{theorem}

\vspace{-0.2cm}

\begin{pf}
    Please refer to~\cite{wang2020understanding}. \hfill$\square$
\end{pf}

Theorem~\ref{th:asm} makes the NCE loss an absolute similarity metric that requires supervision. 
However, note that despite potential ambiguity and heterogeneity for entities in KGs, the aligned pairs should share similar semantic meanings, if not exactly the name. 
Furthermore, the pre-trained word embeddings are known to capture this semantic similarity by projecting similar entities close in the embedding space, which can thus ensure a relatively large $f(x)^Tf(y)$ in Eq.~\ref{eq:nce}, i.e., the ``alignment'' term.  

Therefore, to optimize the NCE loss, the main task is then to optimize the ``uniformity'' term in Eq.~\ref{eq:nce} rather than the ``alignment'' term. 
Considering the boundedness property of $f$, we can instantly draw an unsupervised upper bound of $\mathcal{L}_{\rm ASM}$ by as follows.

\begin{proposition}{\bf Relative similarity metric (RSM).} \label{th:rsm}
For a fixed $\tau > 0$ and encoder $f$ satisfies $\|f(\cdot)\|=1$, we always have the following relative similarity metric plus an absolute deviation controlled by a constant as an upper bound for $\mathcal{L}_{\rm ASM}$:

\beqn{ \footnotesize \label{eq:rsm}
    \begin{aligned}
        \mathcal{L}_{\rm RSM}&= -\frac{1}{\tau} + \expectunder[\substack{
                \{y^-_i\}_{i=1}^M \iidsim p_{\mathsf y}}]
                {\mathop{\log}(e^{1 / \tau} + \sum_ie^{f(x)\T f(y^-_i) / \tau})}\\ 
                & \le \mathcal{L}_{\rm ASM} \le \mathcal{L}_{\rm RSM} + \frac{1}{\tau}\left[1-\underset{(x, y) \sim \distnpos}{\min}\left(f(x)\T f(y)\right)\right].
    \end{aligned}
}

\end{proposition}

\vspace{-0.2cm}

\begin{pf}
    Please refer to Appendix~\ref{sec:proof1}. \hfill$\square$
\end{pf}

By optimizing $\mathcal{L}_{\rm RSM}$, the aligned entities are relatively drawn close by pushing non-aligned ones farther away. 
In other words, if we cannot draw the aligned entities close (e.g., no positive labels), we can instead push those not-aligned ones far away enough.

By analyzing the commonly-used NCE loss for entity alignment, we find that the training can benefit more from pushing those randomly-sampled (negative) pairs far away than pulling aligned (positive) ones close. 
Thus, in \solution, we focus only on attempting to pushing the negatives far away such that we can get rid of the usage of positive data (i.e., labels).

\subsection{Self-Negative Sampling}
\label{sec:sns}

In the analysis above, we demonstrate that to align entities without supervision, the focus of \solution is on sampling negative entity pairs---one from KG $G_x$ and the other from KG $G_y$. 
During negative sampling, without supervision for label-aware negative sampling, it is likely that the underlyingly aligned entity pair is sampled as a negative one, i.e., collision happens. 
Normally, this collision probability can be ignored if a few negatives are sampled; but we discover that a large number of negative samples can be crucial to the success of \solution (Cf. Figure~\ref{fig:size_study}), under which the collision probability is non-negligible (Cf. Table~\ref{tab:ablation}), causing a performance drop by up to 7.7\% relatively.
To mitigate the issue, we propose to sample negatives $x_i^-$ from ${G_x}$ for entity $x\in G_x$, given that we are learning from the uni-space of $G_x$ and $G_y$. 
By doing so, we would avoid the conflict by simply excluding $x$, namely self-negative sampling. 

However, there may be two other issues aroused consequently. 
First, due to the real-world noisy data quality, there may often exist several duplicated $x$ in $G_x$, which could be possibly sampled as negatives. 
Note that this  is also a challenge faced by the supervised  setting, where a few duplicated $y$ may also exist in $G_y$. 
By following the outline of proof in \cite{wang2020understanding},  we show that a certain amount of noise will not influence the convergence of the NCE loss.

\begin{theorem}{\bf (Noisy ASM)} \label{th:nasm}
Let the average duplication factors $\lambda\in\mathbb{N}^+$, $\tau\in\mathbb{R}^+$ be constants. The noisy ASM is denoted as follows and it still converges to the same limit of ASM with the absolute deviation decaying in $\mathcal{O}(M^{-2/3})$.

\beqn{\scriptsize
\label{eq:asm}
\begin{aligned}
\mathcal{L}_{{\rm ASM|}\lambda,\mathsf{x}}(f;\tau,M, p_{\mathsf y})=
\expectunder[\substack{
        (x, y) \sim \distnpos \\
        \{y^-_i\}_{i=1}^M \iidsim p_{\mathsf y}
    }]{- \log \frac{e^{f(x)\T f(y) / \tau}}{\lambda e^{f(x)\T f(y) / \tau} + \sum_i e^{f(x)\T f(y^-_i) / \tau}}}
\end{aligned}
}
\end{theorem}
\begin{pf}
Please refer to Appendix~\ref{sec:proof2}.\hfill$\square$
\end{pf}

The second issue is that by changing the negative samples from $y_i^-\in G_y$ to $x_i^-\in G_x$, we need to confirm whether the $\mathcal{L}_{\rm RSM}$ would still be effective for entity alignment. 
Empirically, for a selected negative sample $y_j^-\in {G_y}$, we can expect there to be some partially similar $x_i^-\in G_x$. 
Since the encoder $f$ is shared for ${G_x}$ and ${G_y}$, the optimization of $f(x_i^-)$ will also contribute to the optimization of $f(y_j^-)$. 
To justify this, we provide the following theorem. 

\begin{theorem}{\bf (Noisy RSM with self-negative sampling)}
Let $\Omega_{\mathsf x}$, $\Omega_{\mathsf y}$ be the spaces of KG triples, respectively,  ${\{x^-_i:\Omega_{\mathsf x}\to\mathbb{R}^n\}}_{i=1}^M$, ${\{y^-_i:\Omega_{\mathsf y}\to\mathbb{R}^n\}}_{i=1}^M$ be i.i.d random variables with distribution $p_{\mathsf x}$, $p_{\mathsf y}$, respectively, 
and $\mathcal{S}^{d-1}$ denote the uni-sphere in $\mathbb{R}^n$. 
If there exists a random variable  $f:\mathbb{R}^n\to\mathcal{S}^{d-1}$ s.t. $f(x_i^-)$ and $f(y_i^-)$ satisfy the same distribution on $\mathcal{S}^{d-1}, 1\le i\le M.$, we then have: 
\beqn{
\lim_{M \rightarrow \infty}|\mathcal{L}_{{\rm RSM|}\lambda,\mathsf{x}}(f;\tau,M,p_{\mathsf x}) - \mathcal{L}_{{\rm RSM|}\lambda,\mathsf{x}}(f;\tau,M,p_{\mathsf y})| = 0.
}
\end{theorem}
\begin{pf}
Please refer to Appendix~\ref{sec:proof3}.\hfill$\square$
\end{pf}

Wang et al.~\cite{wang2020understanding} suggests that under the condition of $p_\mathsf{x}=p_\mathsf{y}$, the encoder $f$ can be attained approximately as the minimizer of the uniform loss.  
Specifically, $f$ follows the uniform distribution on the hypersphere. 
In \solution, the uni-space learning condition ensures the ultimate unified representation for both KGs. 
The initial $p_x$ and $p_y$ are similar but not identical, which indicates that the self-negative sampling is essential. 
However, as the training continues, the encoder will be improved as Theorem~\ref{th:nasm} guarantees to make two KGs more aligned. 
In other words, the entity embeddings of $G_x$ and $G_y$ could be viewed as the samples from one single distribution in a larger space, i.e., $p_\mathsf{x}=p_\mathsf{y}$. 
This in turn allows the existence of $f$ to be more realizable.

\begin{figure}[t]
\centering   
\includegraphics[width=0.99\columnwidth]{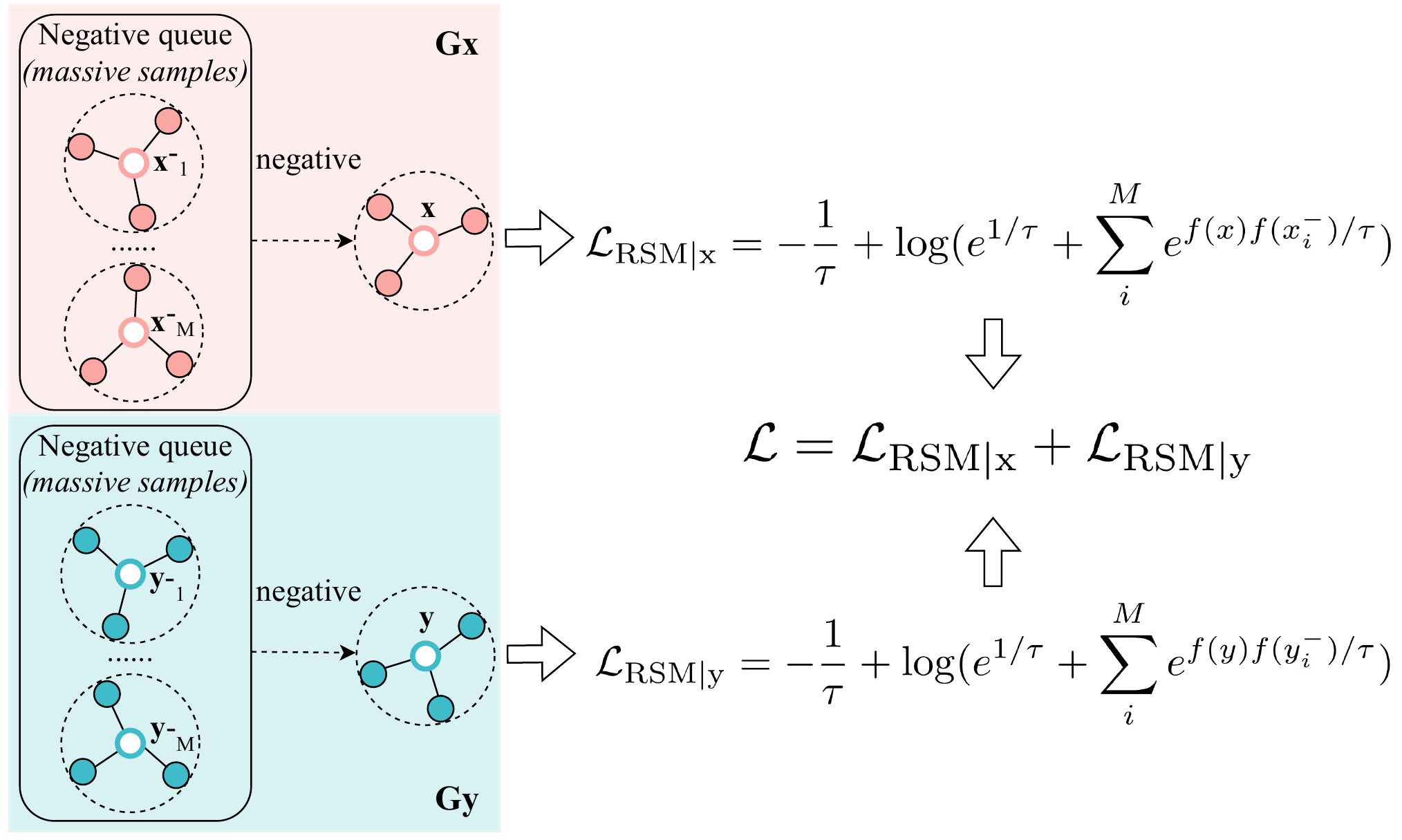}
\caption{The training process of \solution. 
\textmd{It leverages a negative queue for each KG to provide massive negative samples (up to 4k at a time) for calculating the self-supervised contrastive loss.}
}
\label{fig:modelflow}
\vspace{-3mm}
\end{figure}

In practice, we jointly optimize the loss on both ${G_x}$ and ${G_y}$ as follows, which is also illustrated in Figures ~\ref{fig:motivations} (right) and ~\ref{fig:modelflow}. 
\begin{equation} 
    \mathcal{L}=\mathcal{L}_{{\rm RSM|}\lambda,\mathsf{x}}(f;\tau,M,p_{\mathsf x}) + \mathcal{L}_{{\rm RSM|}\lambda,\mathsf{y}}(f;\tau,M,p_{\mathsf y}).
    \label{eq:xy}
\end{equation}

In addition, as the error term of $\mathcal{L}_\lambda(f;\tau,M,p_{\mathsf x})$ decays in $\mathcal{O}(M^{-2/3})$ (Cf. Theorem \ref{th:nasm}), we use a comparatively large number of negative samples to boost the performance.

\subsection{Multiple Negative Queues}
\label{sec:mnq}

Enlarging the number of negative samples can naturally result in additional computational cost, as  encoding massive negative samples on the fly is quite expensive. 
To address this issue, we propose to extend the MoCo technique~\cite{he2020momentum} for \solution. 
In Moco, a negative queue is maintained to store the previously-encoded batches as the encoded negative samples, which host thousands of encoded negative samples at limited cost. 

To adapt to the self-negative sampling strategy in \solution, we practically maintain two negative queues, associating with the two input KGs, respectively. 
An illustrative example is shown in Figure \ref{fig:modelflow}. 
In the beginning, we would not implement the gradient update until one of the queues reaches the predefined length $1+K$ where `$1$' is for the current batch and $K$ is for the number of previous batches used as negative samples. 
Given $|E|$ as the number of entities in a KG, $K$, and the batch size $N$ are constraint by
\begin{equation}
    (1+K)\times N< \min(|E_x|,|E_y|), 
\end{equation}
it is guaranteed that we would not sample out entities in the current batch. 
As a result, the real number of negative samples used for the current batch is $(1 + K)\times N - 1$.

\vpara{Momentum update~\cite{he2020momentum}.} 
The main challenge brought by negative queues is the obsolete encoded samples, especially for those encoded at the early stage of training, during which the model parameters vary drastically. 
Thus, the end-to-end training, which only uses one frequently-updated encoder, may actually harm the training. 
To mitigate this, we adopt the momentum training strategy, which maintains two encoders---the online encoder and the target encoder. 
While the online encoder's parameter $\theta_\mathsf{online}$ is instantly updated with the backpropagation, the target encoder $\theta_\mathsf{target}$ for encoding the current batch and then pushing into the negative queue is asynchronously updated with momentum by:
\begin{equation}
\label{eq:momentum}
    \theta_\mathsf{target} \gets m\cdot\theta_\mathsf{target} + (1-m)\cdot\theta_\mathsf{online}, m\in [0,1)
\end{equation}
A proper momentum is not only important for steady training but may also influence the final performance by avoiding representation collapse (Cf. Figure~\ref{fig:size_study}). We present a series of related hyper-parameter studies in Section~\ref{sec:exp}.

\vpara{Summary.}
We present \solution for self-supervised entity alignment.
\hhy{
Figure \ref{fig:motivations} illustrates that: 1. relative similarity metric (RSM) pushes the non-aligned entities ($y_0^-$, $y_1^-$ and $y_2^-$) of $x$ far enough, instead of directly pulling underlyingly-aligned $y$ close to $x$ (labeled pairs), enabling learning without label supervision; 
2. self-negative sampling samples negative entities for $x$ from ${G_x}$ to avoid sampling the true $y$ as its negative. }
Figure \ref{fig:modelflow} illustrates the training of \solution. 
It leverages existing techniques---embeddings from pre-trained language models and neighborhood aggregator---to initialize entity embeddings into a uni-space.  
The technical contributions of \solution lie in:
\begin{enumerate}
    \item  the design of the self-supervised loss in Eq.~\ref{eq:rsm} enabled by our relative similarity metric (RSM) in KGs; 
    \item  the strategy of self-negative sampling that furthers Eq.~\ref{eq:rsm} into Eq.~\ref{eq:xy} to avoid false-negative samples; 
    \item the extension of MoCo~\cite{he2020momentum} to two negative queues to support an efficient usage of massive negative samples. 
\end{enumerate}

\section{EXPERIMENT} \label{sec:exp}

We evaluate \solution on two widely-acknowledged public benchmarks: DWY100K and DBP15K. DWY100K is a monolingual dataset and DBP15K is a multi-lingual dataset.   


\vpara{DWY100K.} The DWY100K dataset used here is originally built by~\cite{sun2018bootstrapping}. DWY100K consists of two large datasets: DWY100K$_{\text{dbp\_wd}}$ (DBpedia to Wikidata) and DWY100K$_{\text{dbp\_yg}}$ (DBpedia to YAGO3). Each dataset contains 100,000 pairs of aligned entities. However, the entity in the "wd" (Wikidata) part of DWY100K$_{\text{dbp\_wd}}$ are represented by indices (e.g., Q123) instead of URLs containing entity names, and we search their entity names via the Wikidata\footnote{\url{https://pypi.org/project/Wikidata/}} API for python.

\begin{table}[t]
\centering
\renewcommand\tabcolsep{9pt}
  \caption{
    Statistics of DWY100K and DBP15K. \textmd{About the definition of neighbor similarity, please refer to Section~\ref{sec:exp}. ``\#Link" is the number of aligned entity pairs. ``\#Test Link" is the number of aligned pairs for test.}
    }
    \label{data}
    \scalebox{0.8}{
  \begin{tabular}{@{}cccccc@{}}
      \toprule[1.5pt]
      \multirow{2}{*}{Model} & \multicolumn{2}{|c|}{DWY100K}  & \multicolumn{3}{c}{DBP15K} \\ \cmidrule(l){2-6} 
                         & \multicolumn{1}{|c}{dbp\_wd} & \multicolumn{1}{c}{dbp\_yg} & 
                         \multicolumn{1}{|c}{zh\_en} & 
                         \multicolumn{1}{c}{ja\_en} & 
                         \multicolumn{1}{c}{fr\_en} \\ 
    \midrule
        \multicolumn{1}{c|}{\#Link} &
        \multicolumn{1}{c}{99990} &
        \multicolumn{1}{c|}{100000} & 
        \multicolumn{1}{c}{15000} &
        \multicolumn{1}{c}{15000} &
        \multicolumn{1}{c}{15000} \\ 
    \midrule
        \multicolumn{1}{c|}{\#Test Link} &
        \multicolumn{1}{c}{69993} &
        \multicolumn{1}{c}{70000} &
        \multicolumn{1}{|c}{10500} &
        \multicolumn{1}{c}{10500} &
        \multicolumn{1}{c}{10500} \\
    \midrule
        \multicolumn{1}{c|}{\shortstack{neighbor similarity}} &
          \multicolumn{1}{c}{0.633} &
          \multicolumn{1}{c|}{0.777} &
          \multicolumn{1}{c}{0.418} &
          \multicolumn{1}{c}{0.188} &
          \multicolumn{1}{c}{0.182} \\
      \bottomrule[1.5pt]
     \end{tabular}
     }
    \label{tab:stats}
    \vspace{-0.3cm}
\end{table}

\vpara{DBP15K.} The DBP15K dataset is originally built by~\cite{JAPE}\footnote{\url{https://github.com/nju-websoft/JAPE}} and translated into English by~\cite{xu2019cross-lingual}. The DBP15K consists of three cross-lingual datasets: DBP15K$_{\text{zh\_en}}$ (Chinese to English), DBP15K$_{\text{ja\_en}}$ (Japanese to English) and DBP15K$_{\text{fr\_en}}$ (French to English). All three datasets are created from multi-lingual DBpedia, and each contains 15,000 pairs of aligned entities. 
We report results on both original and translated version.


The statistics of DWY100K and DBP15K we use in our work are shown in Table \ref{tab:stats}. 
Beyond basic information, we also present a study on datasets' average (1-hop) neighbor similarity, which is the ratio of aligned neighbors of a pair of aligned entities, indicating how noisy the neighborhood information is.
We observe that DWY100K's neighborhood information is quite useful, while DBP15K's neighborhood information can be very noisy.

\vpara{Experiment Setup.}
We follow the original split of DWY100K \cite{sun2018bootstrapping} and DBP15K \cite{JAPE} which are shown in Table \ref{tab:stats}. 
For \solution, we randomly take out 5\% from the original training set as a dev set for early stopping. 
We use Hit@$k$ $(k=1, 10)$ to evaluate our model's performance as most works do. 
The similarity score is calculated using the $\ell_2$ distance of two entity embeddings. The batch size is set to 64, momentum $m$ is set to $0.9999$, temperature $\tau$ is set to $0.08$, and queue size is set to 64. We use a learning rate of $10^{-6}$ with Adam on a Ubuntu server with NVIDIA V100 GPUs (32G).



\subsection{Results}

In this part, we report the results of \solution and baselines on DWY100K and DBP15K. For all the baselines, we take the reported scores from the corresponding papers, or directly from the tables in BERT-INT~\cite{tang2019bert-int}, CEAFF~\cite{CEAFF} or NAEA~\cite{zhu2019neighborhood}. According to the used proportion of the training labels, we categorize all the models into two types:
\begin{itemize}
    \item Supervised: 100\% of the aligned entity links in the training set is leveraged
    \item Unsupervised \& Self-supervised: 0\% of the training set is leveraged.
\end{itemize}




\begin{table}[t]
	\centering
	\renewcommand\tabcolsep{9pt}
	\renewcommand\arraystretch{0.95}
	\caption{Results on DWY100K. \textmd{Bold results are our best result; underline results are best baseline results.}}
	\scalebox{0.8}{
      \begin{tabular}{@{}cccccc@{}}
          \toprule[1.2pt]
          \multirow{2}{*}{Model} &
            \multicolumn{2}{|c|}{DWY100K$_{\text{dbp\_wd}}$} &
            \multicolumn{2}{|c|}{DWY100K$_{\text{dbp\_yg}}$} &
            \multirow{2}{*}{\makecell[c]{macro\\Hit@1}} \\ 
            \cmidrule(l){2-5} 
           &
            \multicolumn{1}{|c}{Hit@1} &
            \multicolumn{1}{c|}{Hit@10} &
            \multicolumn{1}{c}{Hit@1} &
            \multicolumn{1}{c|}{Hit@10} \\ 
            \midrule
          \multicolumn{6}{c}{Supervised}                     \\ 
          \midrule
          \multicolumn{1}{c|}{MTransE~\cite{MTransE}} &
            \multicolumn{1}{c}{0.281} &
            \multicolumn{1}{c|}{0.520} &
            \multicolumn{1}{c}{0.252} &
            \multicolumn{1}{c|}{0.493} &
            \multicolumn{1}{c}{0.267} \\ \midrule
            \multicolumn{1}{c|}{JAPE~\cite{JAPE}} &
            \multicolumn{1}{c}{0.318} &
            \multicolumn{1}{c|}{0.589} &
            \multicolumn{1}{c}{0.236} &
            \multicolumn{1}{c|}{0.484} &
            \multicolumn{1}{c}{0.277} \\ \midrule
            \multicolumn{1}{c|}{IPTransE~\cite{zhu2017iterative}} &
            \multicolumn{1}{c}{0.349} &
            \multicolumn{1}{c|}{0.638} &
            \multicolumn{1}{c}{0.297} &
            \multicolumn{1}{c|}{0.558} &
            \multicolumn{1}{c}{0.322} \\ \midrule
            \multicolumn{1}{c|}{GCN-Align~\cite{GCN-Align}} &
            \multicolumn{1}{c}{0.477} &
            \multicolumn{1}{c|}{-} &
            \multicolumn{1}{c}{0.601} &
            \multicolumn{1}{c|}{-} &
            \multicolumn{1}{c}{0.539}\\ \midrule
            \multicolumn{1}{c|}{MuGNN~\cite{cao2019multi}} &
            \multicolumn{1}{c}{0.616} &
            \multicolumn{1}{c|}{0.897} &
            \multicolumn{1}{c}{0.741} &
            \multicolumn{1}{c|}{0.937} &
            \multicolumn{1}{c}{0.679}\\ \midrule
            \multicolumn{1}{c|}{RSNs~\cite{guo2019learning}} &
            \multicolumn{1}{c}{0.656} &
            \multicolumn{1}{c|}{-} &
            \multicolumn{1}{c}{0.711} &
            \multicolumn{1}{c|}{-} &
            \multicolumn{1}{c}{0.684}\\ \midrule
            \multicolumn{1}{c|}{BootEA~\cite{sun2018bootstrapping}} &
            \multicolumn{1}{c}{{0.748}} &
            \multicolumn{1}{c|}{0.898} &
            \multicolumn{1}{c}{{0.761}} &
            \multicolumn{1}{c|}{0.894} &
            \multicolumn{1}{c}{0.755}\\ \midrule
            \multicolumn{1}{c|}{NAEA~\cite{zhu2019neighborhood}} &
            \multicolumn{1}{c}{0.767} &
            \multicolumn{1}{c|}{0.918} &
            \multicolumn{1}{c}{0.779} &
            \multicolumn{1}{c|}{0.913} &
            \multicolumn{1}{c}{0.773}\\ \midrule
            
            \multicolumn{1}{c|}{TransEdge~\cite{sun2019transedge}} &
            \multicolumn{1}{c}{0.788} &
            \multicolumn{1}{c|}{0.938} &
            \multicolumn{1}{c}{0.792} &
            \multicolumn{1}{c|}{0.936} &
            \multicolumn{1}{c}{0.790}\\ \midrule
            
            \multicolumn{1}{c|}{RDGCN~\cite{wu2019relation}} &
            \multicolumn{1}{c}{0.902} &
            \multicolumn{1}{c|}{-} &
            \multicolumn{1}{c}{0.864} &
            \multicolumn{1}{c|}{-} &
            \multicolumn{1}{c}{0.883}\\ \midrule
            \multicolumn{1}{c|}{COTSAE~\cite{yang2020cotsae}} &
            \multicolumn{1}{c}{0.927} &
            \multicolumn{1}{c|}{0.979} &
            \multicolumn{1}{c}{0.944} &
            \multicolumn{1}{c|}{0.987} &
            \multicolumn{1}{c}{0.936}\\ \midrule
            \multicolumn{1}{c|}{BERT-INT~\cite{tang2019bert-int}} &
            \multicolumn{1}{c}{0.992} &
            \multicolumn{1}{c|}{-} &
            \multicolumn{1}{c}{0.999} &
            \multicolumn{1}{c|}{-} &
            \multicolumn{1}{c}{0.996}\\ \midrule
            \multicolumn{1}{c|}{CEAFF~\cite{CEAFF}} &
            \multicolumn{1}{c}{\underline{1.000}} &
            \multicolumn{1}{c|}{-} &
            \multicolumn{1}{c}{\underline{1.000}} &
            \multicolumn{1}{c|}{-} &
            \multicolumn{1}{c}{1.000}\\ 
          \midrule
            \multicolumn{6}{c}{Unsupervised \& Self-supervised}              \\ 
            \midrule
            \multicolumn{1}{c|}{MultiKE~\cite{zhang2019multi}} &
            \multicolumn{1}{c}{0.915} &
            \multicolumn{1}{c|}{-} &
            \multicolumn{1}{c}{0.880} &
            \multicolumn{1}{c|}{-} &
            \multicolumn{1}{c}{0.898}\\ \midrule[1.3pt]
          \multicolumn{1}{c|}{\textbf{\solution}} &
            \multicolumn{1}{c}{\textbf{0.983}} &
            \multicolumn{1}{c|}{\textbf{0.998}} &
            \multicolumn{1}{c}{\textbf{1.000}} &
            \multicolumn{1}{c|}{\textbf{1.000}} &
            \multicolumn{1}{c}{0.992}\\ 
          \bottomrule[1.2pt]
         \vspace{-0.8cm}
      \end{tabular}}
	\label{tab:dwy100k}
\end{table}

\vpara{Overall performance on DWY100K.} 
From Table~\ref{tab:dwy100k}, we observe that \solution outperforms all the supervised and unsupervised models except for supervised CEAFF~\cite{CEAFF} and BERT-INT~\cite{tang2019bert-int}. However, without any supervision, \solution only falls behind supervised state-of-the-art CEAFF on DWY100K$_{\text{dbp\_wd}}$ by a minimal margin of 1.2\%. \hhy{The reason why DWY100K$_{\text{dbp\_yg}}$ enables SelfKG to achieve such high accuracy is that the names of its aligned entity pairs are of great similarity respectively, which makes this dataset more easier.} The inspiring result implies that at least for monolingual datasets like DWY100K, supervision is not quite necessary for entity alignment.

\vpara{Overall performance on DBP15K.} 
For the DBP15K dataset, we find that different baselines use different versions of DBP15K in implementation. For example, BERT-INT~\cite{tang2019bert-int} uses the original multi-lingual version built by \cite{JAPE}, while some other methods including RDGCN~\cite{wu2019relation} and DGMC~\cite{fey2020deep} uses machine translation (Google translation) to translate non-English datasets (i.e., zh, ja, fr) of DBP15K into English. If DBP15K is translated, it should not be considered as a multi-lingual setting to some extend. For fair comparison, we report \solution's results on both settings.


\begin{table}[t]
	\centering
    \renewcommand\tabcolsep{3.5pt}
	\caption{Results on DBP15K. \textmd{Methods marked with ``$^*$'' use a translated version of DBP15K~\cite{xu2019cross-lingual}.Bold results are our best result; underline results are best baseline results.}}
	\renewcommand\arraystretch{0.9}
	\scalebox{0.8}{
      \begin{tabular}{@{}cccccccc@{}}
          \toprule[1.2pt]
          \multirow{2}{*}{Model} &
            \multicolumn{2}{|c|}{DBP15K$_{\text{zh\_en}}$} &
            \multicolumn{2}{|c}{DBP15K$_{\text{ja\_en}}$} &
            \multicolumn{2}{|c|}{DBP15K$_{\text{fr\_en}}$} &
            \multirow{2}{*}{\makecell[c]{macro\\Hit@1}} \\ 
            \cmidrule(l){2-7} 
           &
            \multicolumn{1}{|c}{Hit@1} &
            \multicolumn{1}{c|}{Hit@10} &
            \multicolumn{1}{c}{Hit@1} &
            \multicolumn{1}{c|}{Hit@10} &
            \multicolumn{1}{c}{Hit@1} &
            \multicolumn{1}{c|}{Hit@10} &\\ 
            \midrule
          \multicolumn{8}{c}{Supervised}                     \\ 
          \midrule
          \multicolumn{1}{c|}{\begin{tabular}[c]{@{}c@{}}MTransE~\cite{MTransE} \end{tabular}} &
            \multicolumn{1}{c}{0.308} &
            \multicolumn{1}{c|}{0.614} &
            \multicolumn{1}{c}{0.279} &
            \multicolumn{1}{c|}{0.575} &
            \multicolumn{1}{c}{0.244} &
            \multicolumn{1}{c|}{0.556} &
            0.277\\ 
            \midrule
          \multicolumn{1}{c|}{JAPE~\cite{JAPE}} &
            \multicolumn{1}{c}{0.412} &
            \multicolumn{1}{c|}{0.745} &
            \multicolumn{1}{c}{0.363} &
            \multicolumn{1}{c|}{0.685} &
            \multicolumn{1}{c}{0.324} &
            \multicolumn{1}{c|}{0.667} &
            0.366\\
          \midrule
          \multicolumn{1}{c|}{IPTransE~\cite{zhu2017iterative}} &
            \multicolumn{1}{c}{0.406} &
            \multicolumn{1}{c|}{0.735} &
            \multicolumn{1}{c}{0.367} &
            \multicolumn{1}{c|}{0.693} &
            \multicolumn{1}{c}{0.333} &
            \multicolumn{1}{c|}{0.685} &
            0.369\\ 
          \midrule
            \multicolumn{1}{c|}{GCN-Align~\cite{GCN-Align}} &
            \multicolumn{1}{c}{0.413} &
            \multicolumn{1}{c|}{0.744} &
            \multicolumn{1}{c}{0.399} &
            \multicolumn{1}{c|}{0.745} &
            \multicolumn{1}{c}{0.373} &
            \multicolumn{1}{c|}{0.745} &
            0.395\\
            \midrule
          \multicolumn{1}{c|}{SEA~\cite{pei2019semi}} &
            \multicolumn{1}{c}{0.424} &
            \multicolumn{1}{c|}{0.796} &
            \multicolumn{1}{c}{0.385} &
            \multicolumn{1}{c|}{0.783} &
            \multicolumn{1}{c}{0.400} &
            \multicolumn{1}{c|}{0.797} &
            0.403\\
            \midrule
          \multicolumn{1}{c|}{KECG~\cite{li2019semi}} &
            \multicolumn{1}{c}{0.478} &
            \multicolumn{1}{c|}{0.835} &
            \multicolumn{1}{c}{0.490} &
            \multicolumn{1}{c|}{0.844} &
            \multicolumn{1}{c}{0.486} &
            \multicolumn{1}{c|}{0.851} &
            0.485\\ 
            \midrule
          \multicolumn{1}{c|}{MuGNN~\cite{cao2019multi}} &
            \multicolumn{1}{c}{0.494} &
            \multicolumn{1}{c|}{0.844} &
            \multicolumn{1}{c}{0.501} &
            \multicolumn{1}{c|}{0.857} &
            \multicolumn{1}{c}{0.495} &
            \multicolumn{1}{c|}{0.870} &
            0.497\\ 
            \midrule
          \multicolumn{1}{c|}{RSNs~\cite{guo2019learning}} &
            \multicolumn{1}{c}{0.508} &
            \multicolumn{1}{c|}{0.745} &
            \multicolumn{1}{c}{0.507} &
            \multicolumn{1}{c|}{0.737} &
            \multicolumn{1}{c}{0.516} &
            \multicolumn{1}{c|}{0.768} &
            0.510\\ 
            \midrule
          \multicolumn{1}{c|}{AliNet~\cite{DBLP:journals/corr/abs-1911-08936}} &
            \multicolumn{1}{c}{0.539} &
            \multicolumn{1}{c|}{0.826} &
            \multicolumn{1}{c}{0.549} &
            \multicolumn{1}{c|}{0.831} &
            \multicolumn{1}{c}{0.552} &
            \multicolumn{1}{c|}{0.852} &
            0.547\\ 
            \midrule
          \multicolumn{1}{c|}{BootEA~\cite{sun2018bootstrapping}} &
            \multicolumn{1}{c}{0.629} &
            \multicolumn{1}{c|}{0.848} &
            \multicolumn{1}{c}{0.622} &
            \multicolumn{1}{c|}{0.854} &
            \multicolumn{1}{c}{0.653} &
            \multicolumn{1}{c|}{0.874} &
            0.635\\ 
            \midrule
          \multicolumn{1}{c|}{NAEA~\cite{zhu2019neighborhood}} &
            \multicolumn{1}{c}{0.650} &
            \multicolumn{1}{c|}{0.867} &
            \multicolumn{1}{c}{0.641} &
            \multicolumn{1}{c|}{0.873} &
            \multicolumn{1}{c}{0.673} &
            \multicolumn{1}{c|}{0.894} &
            0.655\\ 
            \midrule
          \multicolumn{1}{c|}{MRPEA~\cite{shi2019modeling}} &
            \multicolumn{1}{c}{0.681} &
            \multicolumn{1}{c|}{0.867} &
            \multicolumn{1}{c}{0.655} &
            \multicolumn{1}{c|}{0.859} &
            \multicolumn{1}{c}{0.677} &
            \multicolumn{1}{c|}{0.890} &
            0.671\\ 
            \midrule
          \multicolumn{1}{c|}{JarKA~\cite{chen2020jarka}} &
            \multicolumn{1}{c}{0.706} &
            \multicolumn{1}{c|}{0.878} &
            \multicolumn{1}{c}{0.646} &
            \multicolumn{1}{c|}{0.855} &
            \multicolumn{1}{c}{0.704} &
            \multicolumn{1}{c|}{0.888} &
            0.685\\ 
            \midrule
          \multicolumn{1}{c|}{TransEdge~\cite{sun2019transedge}} &
            \multicolumn{1}{c}{{0.735}} &
            \multicolumn{1}{c|}{{0.919}} &
            \multicolumn{1}{c}{{0.719}} &
            \multicolumn{1}{c|}{{0.932}} &
            \multicolumn{1}{c}{{0.710}} &
            \multicolumn{1}{c|}{{0.941}} &
            0.721\\ 
            \midrule
          \multicolumn{1}{c|}{GM-Align~\cite{xu2019cross-lingual}} &
            \multicolumn{1}{c}{0.679} &
            \multicolumn{1}{c|}{0.785} &
            \multicolumn{1}{c}{0.740} &
            \multicolumn{1}{c|}{0.872} &
            \multicolumn{1}{c}{0.894} &
            \multicolumn{1}{c|}{0.952} &
            0.771\\
            \midrule
          \multicolumn{1}{c|}{{JAPE~\cite{JAPE} $^*$}} &
            \multicolumn{1}{c}{0.731} &
            \multicolumn{1}{c|}{0.904} &
            \multicolumn{1}{c}{{0.828}} &
            \multicolumn{1}{c|}{0.947} &
            \multicolumn{1}{c}{-} &
            \multicolumn{1}{c|}{-} &
            0.780\\
          \midrule
          \multicolumn{1}{c|}{{RDGCN~\cite{wu2019relation} $^*$}} &
            \multicolumn{1}{c}{0.708} &
            \multicolumn{1}{c|}{0.846} &
            \multicolumn{1}{c}{0.767} &
            \multicolumn{1}{c|}{0.895} &
            \multicolumn{1}{c}{0.886} &
            \multicolumn{1}{c|}{0.957} &
            0.787\\
            \midrule
          \multicolumn{1}{c|}{{HGCN~\cite{wu2019jointly} $^*$}} &
            \multicolumn{1}{c}{0.720} &
            \multicolumn{1}{c|}{0.857} &
            \multicolumn{1}{c}{0.766} &
            \multicolumn{1}{c|}{0.897} &
            \multicolumn{1}{c}{0.892} &
            \multicolumn{1}{c|}{0.961} &
            0.793\\
            \midrule
          \multicolumn{1}{c|}{{DGMC~\cite{fey2020deep} $^*$}} &
            \multicolumn{1}{c}{0.801} &
            \multicolumn{1}{c|}{0.875} &
            \multicolumn{1}{c}{0.848} &
            \multicolumn{1}{c|}{0.897} &
            \multicolumn{1}{c}{0.933} &
            \multicolumn{1}{c|}{0.960} &
            0.861\\
            \midrule
           \multicolumn{1}{c|}{RNM~\cite{zhu2020relation}$^*$} &
             \multicolumn{1}{c}{{0.840}} &
             \multicolumn{1}{c|}{{0.919}} &
             \multicolumn{1}{c}{{0.872}} &
             \multicolumn{1}{c|}{{0.944}} &
             \multicolumn{1}{c}{{0.938}} &
             \multicolumn{1}{c|}{{0.954}} &
             0.883\\ \midrule
          \multicolumn{1}{c|}{CEAFF~\cite{CEAFF}} &
            \multicolumn{1}{c}{0.795} &
            \multicolumn{1}{c|}{-} &
            \multicolumn{1}{c}{0.860} &
            \multicolumn{1}{c|}{-} &
            \multicolumn{1}{c}{0.964} &
            \multicolumn{1}{c|}{-} &
            0.873\\ \midrule
          \multicolumn{1}{c|}{HMAN~\cite{yang2019aligning}} &
            \multicolumn{1}{c}{0.871} &
            \multicolumn{1}{c|}{0.987} &
            \multicolumn{1}{c}{0.935} &
            \multicolumn{1}{c|}{0.994} &
            \multicolumn{1}{c}{0.973} &
            \multicolumn{1}{c|}{0.998} &
            0.926\\ 
            \midrule
          \multicolumn{1}{c|}{BERT-INT~\cite{tang2019bert-int}} &
            \multicolumn{1}{c}{\underline{0.968}} &
            \multicolumn{1}{c|}{\underline{0.990}} &
            \multicolumn{1}{c}{\underline{0.964}} &
            \multicolumn{1}{c|}{\underline{0.991}} &
            \multicolumn{1}{c}{\underline{0.992}} &
            \multicolumn{1}{c|}{\underline{0.998}} &
            0.975\\ 
          \midrule
            \multicolumn{8}{c}{Unsupervised \& Self-supervised}              \\ 
            \midrule
          \multicolumn{1}{c|}{MultiKE~\cite{zhang2019multi}} &
            \multicolumn{1}{c}{0.509} &
            \multicolumn{1}{c|}{0.576} &
            \multicolumn{1}{c}{0.393} &
            \multicolumn{1}{c|}{0.489} &
            \multicolumn{1}{c}{0.639} &
            \multicolumn{1}{c|}{0.712} &
            0.514\\
             \midrule[1.3pt]
          \multicolumn{1}{c|}{\textbf{\solution}} &
            \multicolumn{1}{c}{\textbf{0.745}} &
            \multicolumn{1}{c|}{\textbf{0.866}} &
            \multicolumn{1}{c}{\textbf{0.816}} &
            \multicolumn{1}{c|}{\textbf{0.913}} &
            \multicolumn{1}{c}{\textbf{0.957}} &
            \multicolumn{1}{c|}{\textbf{0.992}} &
            0.840\\ 
            \midrule
          \multicolumn{1}{c|}{{{\solution} $^*$}} &
            \multicolumn{1}{c}{0.829} &
            \multicolumn{1}{c|}{0.919} &
            \multicolumn{1}{c}{0.890} &
            \multicolumn{1}{c|}{0.953} &
            \multicolumn{1}{c}{0.959} &
            \multicolumn{1}{c|}{0.992} &
            0.892\\
          \bottomrule[1.2pt]
      \end{tabular}
  }
	\vspace{-5mm}
\end{table}

We observe that \solution beats all previous supervised ones except for HMAN \cite{yang2019aligning}, CEAFF~\cite{CEAFF} and BERT-INT~\cite{tang2019bert-int}. There is a gap between supervised state-of-the-arts and \solution, which indicates that multi-lingual alignment is surely more complicated than the monolingual setting. We also observe a clear gap between different language datasets. DBP15K$_{\text{zh\_en}}$ is the one with the lowest Hit@1, DBP15K$_{\text{ja\_en}}$ is the middle, and DBP15K$_{\text{fr\_en}}$ has the highest score. However, if we recall the neighbor similarity scores presented in Table \ref{tab:stats}, it is the DBP15K$_{\text{zh\_en}}$ that has the highest neighbor similarity. This discovery indicates that the difference in performance can be mostly attributed to challenges brought by multi-lingual setting instead of structural similarities. 



\begin{table*}[tb]
    \caption{
		 Ablation Study of \solution's components and strategies on DWY100K and DBP15K.
	}
	\vspace{-2mm}
    \centering
	\scalebox{1.0}{
    \begin{tabular}{@{}p{4.5cm}<{\centering}cccccccccc@{}}
        \toprule[1.2pt]
        \multirow{2}{*}{Model} & 
        \multicolumn{2}{|c|}{DWY100K$_{\text{dbp\_wd}}$} & 
        \multicolumn{2}{c}{DWY100K$_{\text{dbp\_yg}}$} & 
        \multicolumn{2}{|c|}{DBP15K$_{\text{zh\_en}}$} & 
        \multicolumn{2}{|c}{DBP15K$_{\text{ja\_en}}$}  &  
        \multicolumn{2}{|c}{DBP15K$_{\text{fr\_en}}$} \\ \cmidrule(l){2-11} 
                               
        & \multicolumn{1}{|c}{Hit@1} & 
        \multicolumn{1}{c|}{Hit@10} & 
        \multicolumn{1}{c}{Hit@1} & 
        \multicolumn{1}{c|}{Hit@10} & 
        \multicolumn{1}{c}{Hit@1} & 
        \multicolumn{1}{c|}{Hit@10} & 
        \multicolumn{1}{c}{Hit@1} & 
        \multicolumn{1}{c|}{Hit@10} & 
        \multicolumn{1}{c}{Hit@1} & 
        \multicolumn{1}{c}{Hit@10} \\ \midrule
        
        \multicolumn{1}{l|}{\begin{tabular}[l]{@{}c@{}}\solution\end{tabular}} & 
          \multicolumn{1}{c}{\textbf{0.983}} &
          \multicolumn{1}{c|}{\textbf{0.998}} &
          \multicolumn{1}{c}{\textbf{1.000}} &
          \multicolumn{1}{c|}{\textbf{1.000}} &
          \multicolumn{1}{c}{\textbf{0.745}} &
          \multicolumn{1}{c|}{\textbf{0.866}} &
          \multicolumn{1}{c}{\textbf{0.816}} &
          \multicolumn{1}{c|}{\textbf{0.913}} &
          \multicolumn{1}{c}{\textbf{0.957}} &
          \multicolumn{1}{c}{\textbf{0.992}} \\ 
        \multicolumn{1}{l|}{-w.o. RSM} &
            \multicolumn{1}{c}{0.884} &
            \multicolumn{1}{c|}{0.963} &
            \multicolumn{1}{c}{\textbf{1.000}} &
            \multicolumn{1}{c|}{\textbf{1.000}} &
              \multicolumn{1}{c}{0.670} &
              \multicolumn{1}{c|}{0.813} &
              \multicolumn{1}{c}{0.760} &
              \multicolumn{1}{c|}{0.867} &
              \multicolumn{1}{c}{0.916} &
              \multicolumn{1}{c}{0.987} \\
        \multicolumn{1}{l|}{-w.o. neighbors} &
          \multicolumn{1}{c}{0.887} &
          \multicolumn{1}{c|}{0.987} &
          \multicolumn{1}{c}{\textbf{1.000}} &
          \multicolumn{1}{c|}{\textbf{1.000}} &
          \multicolumn{1}{c}{0.638} &
          \multicolumn{1}{c|}{0.783} &
          \multicolumn{1}{c}{0.732} &
          \multicolumn{1}{c|}{0.849} &
          \multicolumn{1}{c}{0.931} &
          \multicolumn{1}{c}{0.978} \\ 
          \multicolumn{1}{l|}{-w.o. RSM + neighbors} &
            \multicolumn{1}{c}{0.799} &
            \multicolumn{1}{c|}{0.903} &
            \multicolumn{1}{c}{\textbf{1.000}} &
            \multicolumn{1}{c|}{\textbf{1.000}} &
              \multicolumn{1}{c}{0.581} &
              \multicolumn{1}{c|}{0.739} &
              \multicolumn{1}{c}{0.689} &
              \multicolumn{1}{c|}{0.815} &
              \multicolumn{1}{c}{0.899} &
              \multicolumn{1}{c}{0.964} \\ \midrule
          \multicolumn{1}{l|}{-w.o. self negative sampling} &
            \multicolumn{1}{c}{0.918} &
            \multicolumn{1}{c|}{0.978} &
            \multicolumn{1}{c}{\textbf{1.000}} &
            \multicolumn{1}{c|}{\textbf{1.000}} &
              \multicolumn{1}{c}{0.688} &
              \multicolumn{1}{c|}{0.833} &
              \multicolumn{1}{c}{0.773} &
              \multicolumn{1}{c|}{0.882} &
              \multicolumn{1}{c}{0.932} &
              \multicolumn{1}{c}{0.980} \\
        \bottomrule[1.2pt]
        \end{tabular}
        }
    \label{tab:ablation}
\end{table*}

\begin{table*}[tb]
    \caption{
		 Ablation Study on quality of pre-trained uni-space embedding on DWY100K and DBP15K.
	}
	\vspace{-2mm}
    \centering
	\scalebox{1.0}{
    \begin{tabular}{@{}p{4cm}<{\centering}cccccccccc@{}}
        \toprule[1.2pt]
        \multirow{2}{*}{Model} & 
        \multicolumn{2}{|c|}{DWY100K$_{\text{dbp\_wd}}$} & 
        \multicolumn{2}{c}{DWY100K$_{\text{dbp\_yg}}$} & 
        \multicolumn{2}{|c|}{DBP15K$_{\text{zh\_en}}$} & 
        \multicolumn{2}{|c}{DBP15K$_{\text{ja\_en}}$}  &  
        \multicolumn{2}{|c}{DBP15K$_{\text{fr\_en}}$} \\ \cmidrule(l){2-11} 
                               
        & \multicolumn{1}{|c}{Hit@1} & 
        \multicolumn{1}{c|}{Hit@10} & 
        \multicolumn{1}{c}{Hit@1} & 
        \multicolumn{1}{c|}{Hit@10} & 
        \multicolumn{1}{c}{Hit@1} & 
        \multicolumn{1}{c|}{Hit@10} & 
        \multicolumn{1}{c}{Hit@1} & 
        \multicolumn{1}{c|}{Hit@10} & 
        \multicolumn{1}{c}{Hit@1} & 
        \multicolumn{1}{c}{Hit@10} \\ \midrule
        
        \multicolumn{1}{l|}{\begin{tabular}[l]{@{}c@{}}FastText - before \solution training\end{tabular}} & 
          \multicolumn{1}{c}{{0.837}} &
          \multicolumn{1}{c|}{{0.910}} &
          \multicolumn{1}{c}{{0.864}} &
          \multicolumn{1}{c|}{{0.939}}&
          \multicolumn{1}{c}{{0.590}} &
          \multicolumn{1}{c|}{{0.688}} &
          \multicolumn{1}{c}{{0.645}} &
          \multicolumn{1}{c|}{{0.755}} &
          \multicolumn{1}{c}{{0.828}} &
          \multicolumn{1}{c}{{0.898}} \\ 
        \multicolumn{1}{l|}{\qquad\quad\ \ \  - after \solution training} &
            \multicolumn{1}{c}{\textbf{0.921}} &
            \multicolumn{1}{c|}{\textbf{0.986}} &
            \multicolumn{1}{c}{\textbf{0.954}} &
            \multicolumn{1}{c|}{\textbf{0.993}} &
              \multicolumn{1}{c}{\textbf{0.707}} &
              \multicolumn{1}{c|}{\textbf{0.834}} &
              \multicolumn{1}{c}{\textbf{0.755}} &
              \multicolumn{1}{c|}{\textbf{0.865}} &
              \multicolumn{1}{c}{\textbf{0.914}} &
              \multicolumn{1}{c}{\textbf{0.967}} \\ \midrule
              
        \multicolumn{1}{l|}{\begin{tabular}[l]{@{}c@{}}LaBSE - before \solution training\end{tabular}} &
            \multicolumn{1}{c}{0.799} &
            \multicolumn{1}{c|}{0.903} &
            \multicolumn{1}{c}{\textbf{1.000}} &
            \multicolumn{1}{c|}{\textbf{1.000}} &
            \multicolumn{1}{c}{0.581} &
            \multicolumn{1}{c|}{0.739} &
            \multicolumn{1}{c}{0.689} &
            \multicolumn{1}{c|}{0.815} &
            \multicolumn{1}{c}{0.899} &
            \multicolumn{1}{c}{0.964} \\

        \multicolumn{1}{l|}{\qquad\ \ \ \ - after \solution training} &
          \multicolumn{1}{c}{\textbf{0.983}} &
          \multicolumn{1}{c|}{\textbf{0.998}} &
          \multicolumn{1}{c}{\textbf{1.000}} &
          \multicolumn{1}{c|}{\textbf{1.000}} &
          \multicolumn{1}{c}{\textbf{0.745}} &
          \multicolumn{1}{c|}{\textbf{0.866}} &
          \multicolumn{1}{c}{\textbf{0.816}} &
          \multicolumn{1}{c|}{\textbf{0.913}} &
          \multicolumn{1}{c}{\textbf{0.957}} &
          \multicolumn{1}{c}{\textbf{0.992}} \\ 
            
        \bottomrule[1.2pt]
        \end{tabular}
        }
    \label{tab:fasttext}
\end{table*}

\subsection{Ablation Study} \label{sec:ablation}
We conduct extensive ablation studies respectively on DWY100K and DBP15K for \solution. We ablate components regarding the different types of information it brings in. In addition, we conduct studies over some important hyper-parameters using DBP15K$_{\text{zh\_en}}$ dataset as an example.
 
In Table~\ref{tab:ablation}, we present the ablation study for \solution on both DWY100K and DBP15K, including ablation of neighborhood aggregator and ablation of the self-supervised contrastive training objective based on relative similarity metric (RSM) (i.e., use the original encoding outputs from the LaBSE). We first observe that the LaBSE provides rather good initialization. However, merely the LaBSE is not enough. As we can see, on DWY100K, the LaBSE is benefited substantially from our RSM, with an absolute gain over 10\% on DWY100K$_{\text{dbp\_wd}}$ and 5\% on DBP15K. The use of neighborhood aggregator boosts \solution on both DWY100K and DBP15K, which indicates the importance of introducing neighbor information. 

Besides, we test the performance of \solution without self-negative sampling strategy, which means we sample negative entities from the target KG as most baselines do but without labels (which may introduce the true positive ones). The results show that self-negative sampling is necessary for \solution, which brings absolute gains of 2-7\%. While the strategy increase in performance can be partly attributed to avoid of collision, careful readers may think of why possibly-existed duplicated entities does not harm as much as the collision. It can be potentially explained that the entity alignment task evaluates alignment accuracy across different KGs (e.g., $G_x$ and $G_y$) rather than within one KG (e.g., $G_x$). Even though we might sample duplicated entities in $G_x$ and push them away, it might generate only limited influence on their similarities with the target entity $y$ in $G_y$.

\vpara{Impact of the quality of pre-trained uni-space embedding.}
To clarify the influence of different pre-trained word embeddings, we conduct an experiment that replaces the LaBSE embedding we use in \solution with FastText embeddings, which is widely used in baseline methods.

First, comparing FastText results with and without training, the after-training results are consistently higher by 8.5\%-17.2\% than before-training results in Table~\ref{tab:fasttext}. These results also outperform all previous unsupervised baselines, indicating the effectiveness of \solution when being applied to any embedding initialization.

Second, comparing FastText results with LaBSE results, we also confirm that a stronger pre-trained language model like LaBSE will boost \solution's performance compared to FastText word embeddings. This is also the case in baseline methods, such as HMAN~\cite{yang2019aligning} and BERT-INT~\cite{tang2019bert-int}, who leverage multi-lingual BERT as their encoders. Despite better pre-trained embeddings, in our ablation study (Cf. Table~\ref{tab:ablation} and Table~\ref{tab:fasttext}), we show that the "-w.o. RSM + neighbors" (i.e. LaBSE before \solution training) can be significantly improved by 6.4\%-28.2\% with \solution, which demonstrates the usefulness of our method.

\vpara{Impact of relation information and multi-hop structure information.}
To better examine whether relational structural information will help in the self-supervised setting (which might have different results from previous supervised observations), we first conduct experiments on incorporating multi-hop information and then integrate relation information. Table~\ref{tab:multi_hop} shows the results on DBP15K when multi-hop neighbors (more specifically, 20-nearest-neighbor subgraph) are leveraged instead of 1-hop ones. We observe that the performance is actually worse. This is probably because of the heterogeneity of different knowledge graphs and also because the neighbor noises may be amplified in a self-supervised setting. 

\begin{table}[t]
    \caption{
		 Ablation Study of multi-hop structure and relation information on DBP15K.
	}
    \centering
	\label{abdwy2}
    \renewcommand\tabcolsep{3pt}
	\scalebox{0.92}{
    \begin{tabular}{@{}ccccccc@{}}
        \toprule[1.2pt]
        \multirow{2}{*}{Model} & 
        \multicolumn{2}{|c|}{DBP15K$_{\text{zh\_en}}$} & 
        \multicolumn{2}{|c}{DBP15K$_{\text{ja\_en}}$}  &  
        \multicolumn{2}{|c}{DBP15K$_{\text{fr\_en}}$} \\ \cmidrule(l){2-7} 
        &                      
        \multicolumn{1}{|c}{Hit@1} & 
        \multicolumn{1}{c|}{Hit@10} & 
        \multicolumn{1}{c}{Hit@1} & 
        \multicolumn{1}{c|}{Hit@10} & 
        \multicolumn{1}{c}{Hit@1} & 
        \multicolumn{1}{c}{Hit@10} \\ \midrule
        
        \multicolumn{1}{l|}{\begin{tabular}[l]{@{}c@{}}\solution\end{tabular}} & 
          \multicolumn{1}{c}{0.745} &
          \multicolumn{1}{c|}{0.866} &
          \multicolumn{1}{c}{0.816} &
          \multicolumn{1}{c|}{0.913} &
          \multicolumn{1}{c}{0.957} &
          \multicolumn{1}{c}{0.992} \\ \midrule
        \multicolumn{1}{l|}{multi-hop} &
              \multicolumn{1}{c}{0.685} &
              \multicolumn{1}{c|}{0.834} &
              \multicolumn{1}{c}{0.769} &
              \multicolumn{1}{c|}{0.876} &
              \multicolumn{1}{c}{0.936} &
              \multicolumn{1}{c}{0.983} \\ 
        \multicolumn{1}{l|}{with relation} &
              \multicolumn{1}{c}{\textbf{0.750}} &
              \multicolumn{1}{c|}{\textbf{0.876}} &
              \multicolumn{1}{c}{\textbf{0.819}} &
              \multicolumn{1}{c|}{\textbf{0.921}} &
              \multicolumn{1}{c}{\textbf{0.959}} &
              \multicolumn{1}{c}{\textbf{0.994}} \\
        \bottomrule[1.2pt]
        \end{tabular}
        }
    \label{tab:multi_hop}
    \vspace{-4mm}
\end{table}

Based on the 1-hop restriction, as for incorporating relation information, we combine relation name embeddings and their corresponding tail entity name embeddings as the new 1-hop neighbor embeddings. We can see that the results are improved by a slight margin with relation information, which demonstrates that relational information is of a little usefulness.


\begin{figure*}[t]
    \centering
    \setlength{\abovecaptionskip}{2mm}
    \centering
    \includegraphics[width=.98\linewidth]{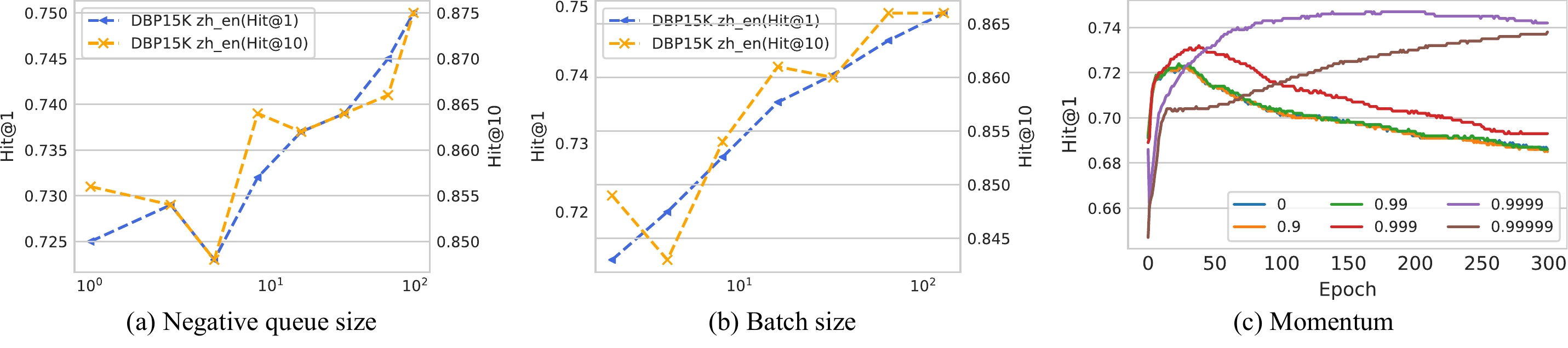}
    \caption{Study on (a) negative queue size, (b) batch size, and (c) momentum on DBP15K$_{\text{zh\_en}}$. (c) presents the test Hit@1 curve throughout the training epochs.}
    \label{fig:size_study}
\end{figure*}

\begin{figure}
    \centering
        \setlength{\abovecaptionskip}{2mm}
        \includegraphics[width=0.9\linewidth]{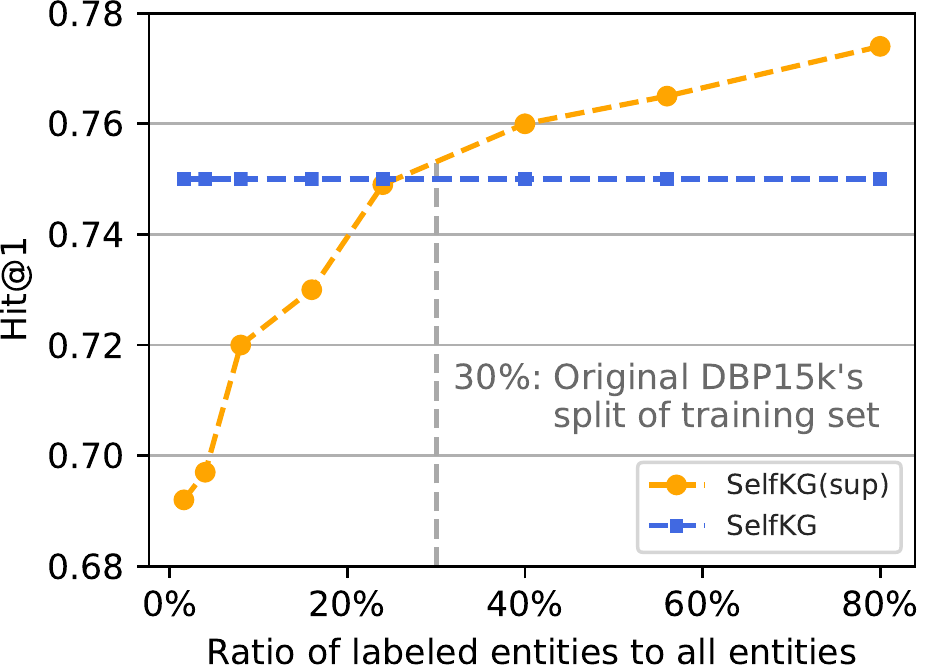}
        \caption{\solution vs. \solution(sup) on DBP15K. \textmd{\solution works well in a low-data resource setting.}}
        \label{fig:sup}
        \vspace{-0.4cm}
\end{figure}

\vpara{Impact of hyper-parameters.} 
The main hyper-parameters in \solution are 1) negative queue size and batch size (which influence the capacity of negative samples), and 2) momentum coefficient $m$ that controls \solution's training stability.

As pointed out in Theorem~\ref{th:asm} and~\ref{th:nasm}, the error term of contrastive loss decays with $\mathcal{O}(M^{-2/3})$, which indicates the importance of enlarging the number of negative samples. Fixing batch size to 64, we change the sizes of the negative queue and derive the curve in Figure~\ref{fig:size_study}. The performance increase is not obvious when queue size is between 10$^0$ and 10$^1$; but as it grows to 10$^2$, the improvement becomes significant. Fixing queue size to 64, along the increase of batch size, the improvement is more stable ranging from 10$^1$ to 10$^2$. 

For momentum coefficient $m$, we discover that a properly large $m$ such as 0.9999 is usually better for \solution. Besides, a proper $m$ is also critical for better training stability (Cf. Figure~\ref{fig:size_study}). A small momentum leads to faster convergence, but also representation collapse and consequent poorer performance. A too-large momentum (e.g., 0.99999) converges too slow.

\subsection{\solution v.s. Supervised \solution}
In practice, we often encounter low-data resource situations where there is very limited supervision. To justify \solution's scalability, we compare self-supervised \solution with its supervised counterpart \solution(sup) on DBP15K$_{\text{zh\_en}}$ across different data resource settings. \solution(sup) follows the conventional supervised entity alignment methods using Absolute Similarity Metric as presented in Eq.~\ref{eq:asm}.

In our preliminary experiment, we find that the original DBP15k's data split (30\% labels for training and 70\% for testing) is not sufficient to present \solution(sup)'s advantage, resulting in a Hit@1 of 0.744 for \solution(sup) and 0.745 for \solution.
So we construct a new split of DBP15K$_{\text{zh\_en}}$ that contains 20\% for testing and 80\% for constructing different sizes of training set. The result is presented in Figure~\ref{fig:sup}, where the horizontal axis indicates the ratio of training labeled entities for \solution(sup) to all entities. We observe that \solution is approximately comparable to \solution(sup) using an amount of 25\% labeled entities, which accords with our observation in the aforementioned preliminary experiment. When using less than 25\% amount of labeled entities, \solution performs much better than \solution(sup), which demonstrates the effectiveness of \solution in low supervised data resource settings.



\section{RELATED WORK}
\vpara{Entity alignment.}
Entity alignment, also named entity resolution, ontology alignment, or schema matching, is a fundamental problem in the knowledge graph community~\cite{zeng2021comprehensive} that has been researched for decades. Before the deep learning era, most approaches focus on designing proper similarity factors and Bayesian-based probability estimation. ~\cite{tang2006using} develops the idea of transforming the alignment into minimizing the risk of decision making. RiMOM~\cite{li2008rimom} proposes a multi-strategy ontology alignment framework, which leverages primary similarity factors with the Cartesian product to align concepts unsupervisedly. ~\cite{li2014rule} argues for rule-based linking and design a rule discovery algorithm. ~\cite{zhang2015cosnet} develops an efficient multi-network linking algorithm based on the factor graph model.

Recently, embedding-based methods have drawn people's attention due to their flexibility and effectiveness. TransE~\cite{bordes2013translating} is the very beginning to introduce the embedding method to represent relational data. ~\cite{MTransE} develops the knowledge graph alignment strategy based on TransE. ~\cite{JAPE} argues for a cross-lingual entity alignment task and constructs the dataset from DBpedia. ~\cite{zhang2018mego2vec} proposes to embed entity ego-network to vectors for the alignment. ~\cite{GCN-Align} introduces the GCN to model both the entity and relation in knowledge graphs to perform the alignment. ~\cite{trisedya2019entity} argues that we can use attributes and structure to supervise each other mutually. BERT-INT~\cite{tang2019bert-int} proposes an interactive entity alignment strategy based on BERT and substantially improves the supervised entity alignment performance on public benchmarks. ~\cite{zhang2019oag} designs heterogeneous graph attention networks to perform large-scale entity linking across the open academic graph.

However, most embedding-based methods nowadays rely heavily on supervised data, hindering their application in real web-scale noisy data. As a prior effort, in~\cite{liu2021oag_know} authors present self-supervised pre-training for concept linking but with downstream supervised classification. In this work, we endeavor to investigate the potential of a completely self-supervised approach without using labels to reduce the cost of entity alignment while improving performance.

\vpara{Self-supervised learning.}
Self-supervised learning~\cite{liu2020self}, which learns the co-occurrence relationships in the data without human supervision, is a data-efficient and powerful machine learning paradigm. We can divide them into two categories: generative and contrastive.

Generative self-supervised learning is often related to pre-training. For instance, BERT~\cite{devlin2018bert}, GPT~\cite{radford2019language}, XLNet~\cite{yang2019xlnet} and so on~\cite{raffel2020exploring,du2021all} develop the field of language model pre-training, which boost the development of natural language processing. The contrastive self-supervised learning is recently proposed by MoCo and SimCLR~\cite{he2020momentum,chen2020simple} in computer vision to conduct successful vision pre-training. The core idea of leveraging the instance discrimination and contrastive loss has been proved to be especially useful for downstream
classification tasks.
Self-supervised learning has also been applied to graph pre-training tasks, such as in GCC~\cite{qiu2020gcc}, the authors pre-train the structural representation of subgraphs using contrastive learning and transfer the model to other graphs. ~\cite{you2020graph} proposes adding augmentations to sampled graphs following SimCLR's strategy to promote graph pre-training performance.

\section{CONCLUSION}
In this work, we re-examine the use and effect of supervision in the entity alignment problem, which targets aligning entities with identical meanings across different knowledge graphs. Based on the three insights we derive---uni-space learning, relative similarity metric, and self-negative sampling, we develop a self-supervised entity alignment algorithm---\solution---to automatically align entities without training labels. The experiments on two widely-used benchmarks DWY100K and DBP15K show that \solution is able to beat or match most of the supervised alignment methods which leverage the 100\% of the training datasets. Our discovery indicates a huge potential to get rid of supervision in the entity alignment problem, and more studies are expected for a deeper understanding of self-supervised learning.


\section*{Acknowledgments}
The work is supported by the NSFC for Distinguished Young Scholar (61825602), NSFC (61836013), and Tsinghua-Bosch Joint ML Center.
Haoyun Hong is supported by Tsinghua University Initiative Scientific
Research Program and DCST Student Academic Training Program. 

\renewcommand\refname{REFERENCES}
\bibliographystyle{ACM-Reference-Format}
\bibliography{ref}

\clearpage

\appendix

\section{Appendix}


\subsection{ Proof to Proposition 1} \label{sec:proof1}

\begin{pf}
Notice that $\frac{x}{x+a}$ is increasing w.r.t $x\in \mathbb{R}, x \geq 0$, where $a\in \mathbb{R}, a > 0$ is a constant. Then we have:
\beqn{\scriptsize
\begin{aligned}
\mathcal{L}_{\rm RSM}
&=\expectunder[\substack{
                \{y^-_i\}_{i=1}^M \iidsim p_{\mathsf y}}]{-\log \frac{e^{\frac{1}{\tau}}}{e^{\frac{1}{\tau}} + \sum_ie^{f(x)\T f(y^-_i) / \tau}}}\\
&\leq \expectunder[\substack{
                (x, y) \sim \distnpos \\
                \{y^-_i\}_{i=1}^M \iidsim p_{\mathsf y}}]{-\log \frac{e^{f(x)\T f(y) / \tau}}{e^{f(x)\T f(y) / \tau} + \sum_ie^{f(x)\T f(y^-_i) / \tau}}}= \mathcal{L}_{\rm ASM}.
\end{aligned}
}

On the other hand,

\begin{equation}{\scriptsize
\begin{aligned}
&\mathcal{L}_{\rm ASM} \leq \expectunder[\substack{(x, y) \sim \distnpos \\
\{y^-_i\}_{i=1}^M \iidsim p_{\mathsf y}}]{-\log \left(\frac{e^{{\min}\left(f(x)\T f(y)\right)/\tau}}{e^{{\min}\left(f(x)\T f(y)\right)/\tau} + \sum_ie^{f(x)\T f(y^-_i)/\tau}}\right)}\\
& \leq  \expectunder[\substack{(x, y) \sim \distnpos \\
\{y^-_i\}_{i=1}^M \iidsim p_{\mathsf y}}]{-\log \left(\frac{e^{{\min}\left(f(x)\T f(y)\right)/\tau}}{e^{\frac{1}{\tau}} + \sum_ie^{f(x)\T f(y^-_i) / \tau}}\right)}\\
& \leq\mathcal{L}_{\rm RSM} + \frac{1}{\tau}\left[1-\underset{(x, y) \sim \distnpos}{\min}\left(f(x)\T f(y)\right)\right]. 
\end{aligned}
}\end{equation}
\hfill $\square$
\end{pf}

\subsection{ Proof to Theorem 2} \label{sec:proof2}

\begin{pf}
We follow the outline of Wang's proof~\cite{wang2020understanding}.

\beqn{\scriptsize
\begin{aligned}
&\lim_{M \rightarrow \infty} [\mathcal{L}_{{\rm ASM|}\lambda,\mathsf{x}}(f;\tau,M, p_{\mathsf y}) - \log M] \\ 
    &= -\frac{1}{\tau}\expectunder[\substack{
        (x, y) \sim \distnpos
    }]{f(x)\T f(y)} 
    \\&+ \lim_{M \rightarrow \infty} \expectunder[\substack{
        (x, y) \sim \distnpos \\
        \{y^-_i\}_{i=1}^M \iidsim p_{\mathsf y}
    }]{\log \left(\frac{\lambda}{M} e^{f(x)\T f(y) / \tau} + \frac{1}{M}\sum_i e^{f(x)\T f(y^-_i) / \tau}\right)}\\
    & = -\frac{1}{\tau}\expectunder[\substack{(x, y) \sim \distnpos}]{f(x)\T f(y)} + \expectunder[\substack{
        x \iidsim p_{\mathsf x}
    }]{\log \expectunder[\substack{
        y^- \iidsim p_{\mathsf y}
    }]{e^{f(x)\T f(y^-) / \tau}}}
\end{aligned}
}

\noindent where the last equality is by the S.L.L.N. (Strong Law of Large Numbers) and the Continuous Mapping Theorem.
\\
\\
\indent The convergence speed is derived as follows, where $\lambda \geq 1$ and $-1 \leq f(x)^Tf(y), f(x)^Tf(y_i^-) \leq 1$.
\\
\\
\indent For one side:
\beqn{\scriptsize
\begin{aligned}
&\mathcal{L}_{{\rm ASM|}\lambda,\mathsf{x}}(f;\tau,M, p_{\mathsf y}) - \log M - \lim_{M \rightarrow \infty}[\mathcal{L}_{{\rm ASM|}\lambda,\mathsf{x}}(f;\tau,M, p_{\mathsf y}) - \log M] \\
    & \leq \expectunder[\substack{
        x \iidsim p_{\mathsf x} \\
        \{y^-_i\}_{i=1}^M \iidsim p_{\mathsf y}
    }]{\log \left(\frac{\lambda}{M} e^{1 / \tau} + \frac{1}{M}\sum_i e^{f(x)\T f(y^-_i) / \tau  }\right)}\\ &- \expectunder[\substack{
        x \iidsim p_{\mathsf x}
    }]{\log \expectunder[\substack{
        y^- \iidsim p_{\mathsf y}
    }]{{e^{f(x)\T f(y^-) / \tau}}}}\\
    &\leq \expectunder[\substack{
        x \iidsim p_{\mathsf x}}]{{\log \expectunder[\substack{
        y^- \iidsim p_{\mathsf y}
    }]{\left(\frac{\lambda}{M} e^{1 / \tau} +  e^{f(x)\T f(y^-) / \tau  }\right)}} - \log \expectunder[\substack{
        y^- \iidsim p_{\mathsf y}
    }]{{e^{f(x)\T f(y^-) / \tau}}}}\\
    & \leq \expectunder[\substack{
        x \iidsim p_{\mathsf x}
    }]{\frac{\lambda}{M} e^{2 / \tau}}\\
    &= \frac{\lambda}{M} e^{2 / \tau},
\end{aligned}
}

\noindent where the second inequality follows the Jensen Inequality based on the the concavity of log.
\\
\\
\indent For the other side:
\beqn{ \scriptsize
\begin{aligned}
&\lim_{M \rightarrow \infty}[\mathcal{L}_{{\rm ASM|}\lambda,\mathsf{x}}(f;\tau,M, p_{\mathsf y}) - \log M] - [\mathcal{L}_{{\rm ASM|}\lambda,\mathsf{x}}(f;\tau,M, p_{\mathsf y}) - \log M]\\
&\leq e^{1/\tau}\expectunder[\substack{
        (x, y) \sim \distnpos \\
        \{y^-_i\}_{i=1}^M \iidsim p_{\mathsf y}
    }]{\left| \expectunder[\substack{
        y^- \iidsim p_{\mathsf y}
    }]{e^{f(x)\T f(y^-) / \tau}} - \left(\frac{\lambda}{M} e^{f(x)\T f(y) / \tau} + \frac{1}{M}\sum_i e^{f(x)\T f(y^-_i) / \tau  }\right)\right|}\\
& \leq \frac{\lambda}{M} e^{2 / \tau} + e^{1/\tau} \expectunder[\substack{
        (x, y) \sim \distnpos \\
        \{y^-_i\}_{i=1}^M \iidsim p_{\mathsf y}
    }]{\left| \expectunder[\substack{
        y^- \iidsim p_{\mathsf y}
    }]{e^{f(x)\T f(y^-) / \tau}} - \frac{1}{M}\sum_i e^{f(x)\T f(y^-_i) / \tau  } \right|}\\
& \leq \frac{\lambda}{M} e^{2 / \tau} + \frac{5}{4}M^{-\frac{2}{3}}e^{\frac{1}{\tau}}\left(e^{\frac{1}{\tau}} - e^{-\frac{1}{\tau}}\right),
\end{aligned}
}

\noindent where the first inequality follows an application of Lagrange's mean-value theorem, and the last inequality follows the bound from Chebychev’s inequality, which can refer to ~\cite{wang2020understanding}.
\\
\\
Therefore, The noisy ASM still converges to the same limit of ASM with absolute deviation decaying in $\mathcal{O}(M^{-2/3})$, combing the derivations of both sides above.\hfill$\square$
\end{pf}

\subsection{ Proof to Theorem 3} \label{sec:proof3}

Let $\Omega_{\mathsf x}, \Omega_{\mathsf y}$ be the space of knowledge graph triplets, $n\in\mathbb{N}$. Let ${\{x^-_i:\Omega_{\mathsf x}\to\mathbb{R}^n\}}_{i=1}^M$, ${\{y^-_i:\Omega_{\mathsf y}\to\mathbb{R}^n\}}_{i=1}^M$ be i.i.d random variables with distribution $p_{\mathsf x}, p_{\mathsf y}$. $\mathcal{S}^{d-1}$ denotes the uni-sphere in $\mathbb{R}^n$. If there exists a random variable  $f:\mathbb{R}^n\to\mathcal{S}^{d-1}$ s.t. $f(x_i^-),f(y_i^-)$ satisfy the same distribution on $\mathcal{S}^{d-1}, 1\le i\le M.$, then we have

\beqn{\scriptsize
\lim_{M \rightarrow \infty}\left|\mathcal{L}_{{\rm RSM|}\lambda,\mathsf{x}}(f;\tau,M,p_{\mathsf x}) - \mathcal{L}_{{\rm RSM|}\lambda,\mathsf{x}}(f;\tau,M,p_{\mathsf y})\right| = 0.
}

\begin{pf}

\begin{equation}{\scriptsize
\begin{aligned}
&\left|\mathcal{L}_{{\rm RSM|}\lambda,\mathsf{x}}(f;\tau,M,p_{\mathsf x}) - \mathcal{L}_{{\rm RSM|}\lambda,\mathsf{x}}(f;\tau,M,p_{\mathsf y})\right|\\
&=\left| \expectunder[\substack{
        \{x^-_i\}_{i=1}^M \iidsim p_{\mathsf x}
    }]{- \log \left(\frac{e^{\frac{1}{\tau}}}{\lambda e^{\frac{1}{\tau}} + \sum_i e^{f(x)\T f(x^-_i) / \tau}}\right)} \right. \\ & \quad \left. - \expectunder[\substack{
        \{y^-_i\}_{i=1}^M \iidsim p_{\mathsf y}
    }]{- \log \left(\frac{e^{\frac{1}{\tau}}}{\lambda e^{\frac{1}{\tau}} + \sum_i e^{f(x)\T f(y^-_i) / \tau}}\right)} \vphantom{\int_1^2} \right|\\
&= \left| \expectunder[\substack{
        \{x^-_i\}_{i=1}^M \iidsim p_{\mathsf x}\\
        \{y^-_i\}_{i=1}^M \iidsim p_{\mathsf y}
    }]{\log \left (\frac{\lambda e^{\frac{1}{\tau}} + \sum_i e^{f(x)\T f(x^-_i) / \tau}}{\lambda e^{\frac{1}{\tau}} + \sum_i e^{f(x)\T f(y^-_i) / \tau}}\right)}\right|\\
& \leq  \expectunder[\substack{
        \{x^-_i\}_{i=1}^M \iidsim p_{\mathsf x}\\
        \{y^-_i\}_{i=1}^M \iidsim p_{\mathsf y}
    }]{\left | \log \left(\frac{\lambda e^{\frac{1}{\tau}} + \sum_i e^{f(x)\T f(x^-_i) / \tau}}{\lambda e^{\frac{1}{\tau}} + \sum_i e^{f(x)\T f(y^-_i) / \tau}}\right) \right |} \\
&= \expectunder[\substack{
        \{x^-_i\}_{i=1}^M \iidsim p_{\mathsf x}\\
        \{y^-_i\}_{i=1}^M \iidsim p_{\mathsf y}
    }]{\left | \log \left( 1 + \frac{ \sum_i e^{f(x)\T f(x^-_i) / \tau} - \sum_i e^{f(x)\T f(y^-_i) / \tau}}{\lambda e^{\frac{1}{\tau}} + \sum_i e^{f(x)\T f(y^-_i) / \tau}}\right) \right |}.
\end{aligned}
}\end{equation}

Let $S = \frac{\lambda e^{\frac{1}{\tau}} + \sum_i e^{f(x)\T f(x^-_i) / \tau}}{\lambda e^{\frac{1}{\tau}} + \sum_i e^{f(x)\T f(y^-_i) / \tau}}$, then

\beqn{\scriptsize
S \geq \frac{\lambda e^{\frac{1}{\tau}} + Me^{-\frac{1}{\tau}}}{\lambda e^{\frac{1}{\tau}} + Me^{\frac{1}{\tau}}} \geq e^{-\frac{2}{\tau}}.
}

Let $T = \frac{ \sum_i e^{f(x)\T f(x^-_i) / \tau} - \sum_i e^{f(x)\T f(y^-_i) / \tau}}{\lambda e^{\frac{1}{\tau}} + \sum_i e^{f(x)\T f(y^-_i) / \tau}}$, therefore

\beqn{\scriptsize
\left| T \right | = \frac{ \frac{1}{M} \left|\sum_i e^{f(x)\T f(x^-_i) / \tau} - \sum_i e^{f(x)\T f(y^-_i) / \tau}\right|}{\frac{1}{M} \left(\lambda e^{\frac{1}{\tau}} + \sum_i e^{f(x)\T f(y^-_i) / \tau}\right)}\leq \frac{e^{\frac{1}{\tau}}-e^{-{\frac{1}{\tau}}}}{\frac{\lambda}{M}e^{\frac{1}{\tau}}+e^{-\frac{1}{\tau}}}.
}

Then $S = 1+T$, therefore 

\beqn{\scriptsize
S \leq 1+ \frac{e^{\frac{1}{\tau}}-e^{-{\frac{1}{\tau}}}}{\frac{\lambda}{M}e^{\frac{1}{\tau}}+e^{-\frac{1}{\tau}}}  <  1+ \frac{e^{\frac{1}{\tau}}-e^{-{\frac{1}{\tau}}}}{e^{-\frac{1}{\tau}}} = 1+e^{\frac{2}{\tau}} - 1 = e^{\frac{2}{\tau}}.
}

Therefore, $\left|\log S\right|<\frac{2}{\tau}$.

By the S.L.L.N., $\lim_{M \rightarrow \infty} T = 0$, therefore, $\lim_{M \rightarrow \infty} \log S = 0$.

Because $\left | \log S \right |$ is bounded, with Dominated Covergence Theorem, the sign of mathematical expectation (i.e. integral) can be exchanged with the sign of limit:

\beqn{\scriptsize
\begin{aligned}
& \lim_{M \rightarrow \infty}\left|\mathcal{L}_{{\rm RSM|}\lambda,\mathsf{x}}(f;\tau,M,p_{\mathsf x}) - \mathcal{L}_{{\rm RSM|}\lambda,\mathsf{x}}(f;\tau,M,p_{\mathsf y})\right| \\ 
&\leq \lim_{M \rightarrow \infty}{\expectunder(\substack{
        \{x^-_i\}_{i=1}^M \iidsim p_{\mathsf x}\\
        \{y^-_i\}_{i=1}^M \iidsim p_{\mathsf y}
    }){\left | \log S \right|}} \\ 
    &= \expectunder[\substack{
        \{x^-_i\}_{i=1}^M \iidsim p_{\mathsf x}\\
        \{y^-_i\}_{i=1}^M \iidsim p_{\mathsf y}
    }]{\lim_{M \rightarrow \infty}{\left | \log S \right|}}\\& = 0.
\end{aligned}
}
\hfill $\square$
\end{pf}

\hide{Wang et al.\cite{wang2020understanding} suggest that under the condition of $p_\mathsf{x}=p_\mathsf{y}$, the encoder $f$ can be attained approximately as the minimizer of the uniform loss.  Specifically, $f$ follows the uniform distribution on the hypersphere. In our framework, the uni-space learning condition ensures us to obtain unified
representation for both KGs; in other words, entity embeddings of KG$_\mathrm{x}$ and KG$_\mathrm{y}$ could be viewed as samples from one single distribution in a larger space, i.e., $p_\mathsf{x}=p_\mathsf{y}$. This in turn allows the existence of $f$ to be more realizable.
To further this thinking, one may obtain a larger number $M$ of negative samples by randomly generating word vectors and neighbors. }

\vspace{-5mm}
\subsection{Details on Implementation} \label{sec:exp_details}

\subsubsection{Dataset}

For both DWY100K\footnote{Can be downloaded from https://github.com/nju-websoft/BootEA} and DBP15K\footnote{Can be downloaded from https://github.com/syxu828/Crosslingula-KG-Matching} datasets we used, we do simple data processing on the original datasets built in BootEA ~\cite{sun2018bootstrapping} and JAPE ~\cite{JAPE} respectively. The process of data processing is as follows:

Firstly, we remove the redundant prefixes of the URLs representing the entities, leaving the meaningful entity names at the end. For example, in DBP15K$_{\text{zh\_en}}$ dataset, there is an entity represented by "http://dbpedia.org/resource/2012\_Summer\_Olympics". We remove the substring in front of "2012\_Summer\_Olympics" to remove the useless part. Then we replace the underscores used to connect words in the entity names with spaces, so that the entities can be represented by their original entity names. In addition, we replace the indices that represent entities in DWY100K$_{\text{dbp\_wd}}$ (e.g., Q123) with strings of entity names. The purpose of this step is to make the entity names as original as possible to let our model better extract the character-level information and the semantic-level information with useful data. Then, we need to map every entity to a unique index in every pair of KGs respectively. The pairs of KGs are the subdatasets of DWY100K and DBP15K: DWY100K$_{\text{dbp\_wd}}$, DWY100K$_{\text{dbp\_yg}}$, DBP15K$_{\text{zh\_en}}$, DBP15K$_{\text{ja\_en}}$ and DBP15K$_{\text{fr\_en}}$. We use DBP15K dataset provided in~\cite{xu2019cross-lingual} and the DWY100K dataset provided in \cite{JAPE} as our original dataset and follow the indices they created in our experiments since they have already done this processing step.

As for obtaining 1-hop neighbors, we treat the KGs as undirected graphs, that means we use the relational triples in the datasets to find all the entities connected to an entity regardless of the direction of the connection.

Finally, we reconstruct Dataset and use DataLoader of Pytorch's torch.utils.data package to packet our data and create batches. Because we do not use any labels in our model for training, we set the indices of the entities as the y data which is usually used to contain the labels in Dataset package. As for the x data which is the training data in Dataset package, we set the entity names of the center entities and the corresponding neighbors with the adjacency matrix of the center entities as the x data.

\subsubsection{Implementation Notes}

Our model is implemented using Python package Pytorch 1.7.1.\footnote{More details can be found in our code \url{https://github.com/THUDM/SelfKG}}
The experiments were conducted on a GNU/Linux server with 8 Tesla V100 SXM2 GPU and 32G GPU RAM mainly, and also 56 Intel(R) Xeon(R) Gold 5120 CPU(2.20GHz), 500G RAM.





For both experiments on DWY100K and DBP15K, we randomly select 5\% links from the training set in the original datasets as our validation set and evaluate our model's performance both on the validation set and the testing set. We stop the training progress once our model reaches the best performance on the validation set and record Hit@1 and Hit@10 results on the testing set.

\subsubsection{Similarity Search}

In order to evaluate our model on the validation set and the test set efficiently, we apply Faiss\footnote{\url{https://github.com/facebookresearch/faiss}}, a library for efficient similarity search.

In the evaluation period, we apply the IndexFlatL2 as indexer, which is based on $\ell_2$ distance. Once the indices are built, via the kd-tree algorithm used in Faiss, the top 1 and top 10 closest entities in the target KG of every entity in the source KG can be found efficiently.

\subsection{Runtime}

\hhy{On the time efficiency of using large number negative samples in \solution, by leveraging multiple negative queues with Moco~\cite{he2020momentum}, the running time of \solution is significantly reduced even when the sample size is large, making it similar to the common negative sampling method adopted in state-of-the-art baseline methods. Details are discussed in Section~\ref{sec:mnq}.}

\subsection{Limitations}

\hhy{There are mainly two limitations in \solution. Firstly, \solution requires good embeddings to ensure the unified representation for both KGs. As we clarified in Section \ref{sec:ablation}, we confirm that a better pre-trained language model like LaBSE will boost the performance of \solution. This issue is also commonly faced by other embedding-based entity alignment methods. Secondly, \solution still underperforms some supervised state-of-the-art methods. Some of the supervised methods such as BERT-INT\cite{tang2019bert-int} can reach almost an accuracy of 100\% on both DBP15K and DWY100K, which outperforms our self-supervised solution. The gap is expected since supervision does provide much useful information for the alignment task. The ultimate goal of self-supervised methods is to match or even beat supervised methods.}

\end{document}